\documentclass[sigconf]{acmart}
\AtBeginDocument{%
  }

\setcopyright{none}
\renewcommand\footnotetextcopyrightpermission[1]{} 

\newcommand{\prompt}[2]{
    \vspace{0.1in}
    \fbox{
        \parbox{0.9\columnwidth}{
            \underline{\textbf{{#1}:}}
            \newline 
            {#2}
            }
        }
    \vspace{0.1in}
}

\begin{document}

\title{Accuracy and Political Bias of News Source Credibility Ratings by Large Language Models}

\author{Kai-Cheng Yang}
\affiliation{%
  \institution{Northeastern University}
  \city{Boston}
  \state{Massachusetts}
  \country{USA}
}
\email{yang3kc@gmail.com}

\author{Filippo Menczer}
\affiliation{%
  \institution{Indiana University}
  \city{Bloomington}
  \state{Indiana}
  \country{USA}
}

\renewcommand{\shortauthors}{Yang and Menczer}

\begin{abstract}
Search engines increasingly leverage large language models (LLMs) to generate direct answers, and AI chatbots now access the Internet for fresh data.
As information curators for billions of users, LLMs must assess the accuracy and reliability of different sources.
This paper audits nine widely used LLMs from three leading providers---OpenAI, Google, and Meta---to evaluate their ability to discern credible and high-quality information sources from low-credibility ones.
We find that while LLMs can rate most tested news outlets, larger models more frequently refuse to provide ratings due to insufficient information, whereas smaller models are more prone to making errors in their ratings.
For sources where ratings are provided, LLMs exhibit a high level of agreement among themselves (average Spearman's $\rho = 0.79$), but their ratings align only moderately with human expert evaluations (average $\rho = 0.50$).
Analyzing news sources with different political leanings in the US, we observe a liberal bias in credibility ratings yielded by all LLMs in default configurations.
Additionally, assigning partisan roles to LLMs consistently induces strong politically congruent bias in their ratings.
These findings have important implications for the use of LLMs in curating news and political information.
\end{abstract}





\maketitle

\section{Introduction}

Large language models (LLMs) are being integrated into our information ecosystems, transforming how people seek and consume information.
One significant trend is the emergence of AI-augmented search, where an LLM layer is added to traditional search engines to provide direct answers based on search results~\cite{xiong2024searchengineservicesmeet}.
Major platforms like Google\footnote{blog.google/products/search/generative-ai-google-search-may-2024} and Microsoft\footnote{blogs.microsoft.com/blog/2023/02/07/reinventing-search-with-a-new-ai-powered-microsoft-bing-and-edge-your-copilot-for-the-web} have implemented this feature, and newer products like Perplexity AI and You.com have gained substantial user bases and investments.\footnote{techcrunch.com/2024/01/04/ai-powered-search-engine-perplexity-ai-now-valued-at-520m-raises-70m}
Additionally, AI chatbots are now often connected to the Internet, allowing them to fetch up-to-date information not included in their training data and ground their responses in real-time~\cite{vu2023freshllmsrefreshinglargelanguage}.
In these systems, LLMs act as information curators, determining what content is shown to billions of users.
Recent studies suggest that such integration lowers the information access barrier~\cite{wu2020providing} and enables users to perform complex tasks more quickly~\cite{spatharioti2023comparingtraditionalllmbasedsearch,suri2024usegenerativesearchengines}, indicating potential for mainstream adoption.

Despite such advantages, the additional LLM components could introduce new problems~\cite{memon2024searchenginespostchatgptgenerative} because language models face critical technical challenges, such as hallucinations~\cite{ji2023survey}, and exhibit various biases~\cite{gallegos2024bias}.
Recent audits of popular AI search engines reveal that their results often contain unsupported claims~\cite{liu2023evaluatingverifiabilitygenerativesearch} and biases depending on the queries~\cite{li2024generativeaisearchengines}.
Furthermore, experiments conducted by \citet{sharma2024generative} demonstrate that users may engage with more biased information when interacting with AI search engines, and LLMs with predefined opinions can exacerbate this bias.
Beyond these analyses, our understanding of the potential issues of the new LLM layer remains limited.

This study investigates whether LLMs exhibit errors and political biases when evaluating the credibility of information sources, as these issues could compromise their effectiveness as information curators.
We conduct comprehensive experiments to audit nine prominent LLMs from three leading providers: OpenAI, Meta, and Google.
Our methodology involves instructing these models to evaluate and rate over 7,000 websites that represent major news sources across the Internet.
We assess the accuracy of these LLM-generated ratings by comparing them with evaluations provided by human experts.

After reviewing the related work and describing our methodology, we present the results of our experiments.
For most tested news sources, we find that the LLMs can provide ratings as instructed.
However, larger models more frequently refuse to rate less popular sources due to insufficient knowledge, while smaller models are more prone to make errors.
The LLM ratings show a high level of agreement among themselves, even though they are trained by different providers.
However, their ratings only moderately correlate with those of human experts.
Focusing on news sources with clear political leanings in the US, we find that assigning partisan roles (e.g., Democrat and Republican) to LLMs consistently biases their ratings in favor of sources with congruent leanings.
Additionally, LLMs exhibit a liberal bias in their default configurations.

Our findings suggest that while LLMs have the potential to judge the credibility of information sources, state-of-the-art models from different companies exhibit common problems.
Their lack of knowledge about unpopular information sources presents challenges in dealing with data voids~\cite{boyd2018data,aslett2024online}.
Furthermore, inaccurate LLM ratings due to issues like errors and biases could inadvertently amplify misinformation while suppressing high-credibility sources.
When handling political information, the inherent partisan biases in LLMs could exacerbate echo chambers and polarization~\cite{cinelli2021echo,conover2021political}.
Therefore, we caution against relying solely on LLMs for information curation and call for further evaluations and improvements to ensure their reliability and accuracy.

\section{Related Work}

Our study follows a long line of scholarship aimed at understanding how online platforms affect user information diets~\cite{rahwan2019machine} and their contributions to issues such as the proliferation of misinformation~\cite{lazer2018science}, echo chambers~\cite{cinelli2021echo}, and polarization~\cite{conover2021political}.
As prominent platforms for information-seeking, search engines have been scrutinized by many researchers evaluating if their ranking algorithms favor certain news sources~\cite{robertson2018auditingpartisan,trielli2019search}.
Social media platforms are another focus of attention.
Previous studies address topics such as political biases and the amplification of malicious content in various platforms, including Twitter/X~\cite{chen2021neutral}, Facebook~\cite{bakshy2015exposure}, and YouTube~\cite{hosseinmardi2021examing}.

The emergence of LLMs such as ChatGPT has initiated a new line of research auditing these models to ensure they are deployed in ways that are ethical, safe, and accountable~\cite{mokander2023auditing}.
Recent studies show that LLMs can perpetuate discrimination due to their inherent biases~\cite{bai2024measuringimplicitbiasexplicitly,salinas2023imracistbutdiscovering} and stereotypes~\cite{abid2021persistent,cheng2023markedpersonasusingnatural}.
In addition, LLMs have often been found to exhibit a liberal bias in the political context~\cite{hartmann2023politicalideologyconversationalai,santurkar2023whose}.

As LLMs are increasingly integrated into online platforms, there is a growing need for research to evaluate the potential risks associated with this use.
Several recent studies have begun to address this topic by auditing AI search engines such as Bing Chat, Google Bard, and Perplexity AI.
By querying these tools with various search phrases, \citet{liu2023evaluatingverifiabilitygenerativesearch} show that the responses often contain unsupported claims~\cite{liu2023evaluatingverifiabilitygenerativesearch}; \citet{li2024generativeaisearchengines} identify sentiment and geographic biases, and \citet{urman2023silence} reveal significant disparities across different services when dealing with political information.
Through experiments in which participants interact with AI search engines in a lab environment, \citet{sharma2024generative} find that users tend to engage with more biased information, and opinionated LLMs can exacerbate this bias.
Despite these research efforts, our understanding of the impact of LLMs as information curators remains limited.

Our study evaluates the capability of LLMs to assess the credibility of information sources.
Traditionally, this assessment has been conducted by human experts.
Organizations such as NewsGuard\footnote{newsguardtech.com/ratings/rating-process-criteria} and MBFC\footnote{mediabiasfactcheck.com/methodology} employ teams of professional fact-checkers and media literacy experts who systematically evaluate and rate the credibility of diverse information sources.
Recent research has started to explore automated approaches to this challenge.
\citet{carragher2024detection} investigates the application of web graphs for automated detection of unreliable domains, while \citet{williams2024bridgingsocialmediasearch} demonstrates enhanced performance by combining signals from web graphs and social media.
Unlike these studies, we instruct LLMs to finish the tasks in a zero-shot manner in our experiments without training or providing them with additional data.

\section{Methods}

\subsection{Model Selection}

We audit nine state-of-the-art models from three leading providers.
These models have been deployed through various services and systems, which directly or indirectly interact with billions of users across the world every day.

For OpenAI, we test GPT-4o mini (gpt-4o-mini-2024-07-18), GPT-4o (gpt-4o-2024-05-13), and GPT-4 (gpt-4-turbo-2024-04-09).
Although OpenAI has not released the number of parameters used by these models, the price and response time suggest that GPT-4o mini is the smallest and GPT-4 is the largest.
Additionally, we test the o1-mini (o1-mini-2024-09-12) model, which is designed for complex reasoning tasks by explicitly incorporating chain-of-thought into its inference process~\cite{openai2024openaio1card}.
Such a design improves o1-mini's performance on certain tasks, but its number of parameters is not comparable to the other models.
These OpenAI models are available to users through ChatGPT and serve as the backend for AI search engines like Bing.
In our experiment, we query OpenAI's API endpoints directly.

For Meta, we test Llama 3.1 models with 8B, 70B, and 405B parameters~\cite{dubey2024llama3herdmodels}.
Meta has incorporated these models into its products, such as WhatsApp and Facebook, allowing users to interact with them directly.
Due to their open-weight nature, Llama models have also been adopted in many third-party services and applications.
Here, we query the Llama 3.1 models hosted by \url{together.ai} through its API endpoints.

For Google, we test Gemini 1.5 Flash (gemini-1.5-flash-001) and Pro (gemini-1.0-pro-001), which are available to users through the Gemini App.
Google also leverages these models to provide AI summaries in its search results.
We query Google API endpoints directly and disable all safety filters.

\subsection{Prompts}

For all models, we set the temperature parameter to zero and use the same prompts.
The system prompt reads as follows:

\prompt{System prompt}{
You are an assistant to determine the credibility of websites.
}

Then, we provide the model with the task instructions:

\prompt{Task instruction}{
Rate the website's credibility: \{domain\}, on a scale between 0 and 1, where 0 means very low credibility and 1 means very high credibility. The assistant returns a rating of $-1$ when the assistant has no knowledge of the website; otherwise, the assistant should provide the best estimation. In addition to the rating, the assistant should also provide a short explanation.
}

In each query, \{domain\} is replaced with the news website of interest.
Here, we only provide the LLMs with the domain names to test if the LLMs have knowledge of the news sources encoded.
Note that the complete instructions also include specifications for the output; see the Appendix for details.

The prompt above tests the default configuration of LLMs.
To further investigate how easily the responses of LLMs can be steered, we also test the impact of assigning partisan roles to LLMs in the prompts.
Specifically, we append the following instructions to the system prompt:

\prompt{Partisan role assignment}{
You identify as \{role\} on the US political spectrum.
}

In the experiments, \{role\} is replaced with the one of ``a Democrat,'' ``an Independent,'' or ``a Republican'' to induce different political biases.

\subsection{News Outlet Credibility Ratings}

In this study, we adopt ratings of news source credibility compiled by \citet{lin2023high}.
The authors analyze the news ratings from six sources, including NewsGuard,\footnote{newsguardtech.com/ratings/rating-process-criteria} \url{adfontesmedia.com}, Iffy index of unreliable sources,\footnote{iffy.news} MBFC,\footnote{mediabiasfactcheck.com/methodology} and two lists compiled by professional fact-checkers and researchers~\cite{pennycook2019fighting,lasser2022social}.
The comparison of these ratings reveals a high level of consistency.
Using an ensemble method, they generate an aggregate list that contains credibility ratings for 11,520 websites.

\subsection{Website Popularity Ranking}

Our examination of the news source list from \citet{lin2023high} reveals that it contains websites that are no longer active so testing with them is not meaningful.
To remove these websites, we leverage the Tranco list that measures website popularity~\cite{pochat2018tranco}.
The Tranco list combines the website ranking information from multiple sources, including Alexa\footnote{alexa.com/topsites} and Cisco Umbrella.\footnote{umbrella-static.s3-us-west-1.amazonaws.com/index.html}
It is updated on a routine basis, and for this study, we use the snapshot from September 2022.
This snapshot contains the ranks of the top 5.5 million websites worldwide.
Its intersection with the list from \citet{lin2023high} contains 7,523 websites.
In the following, we refer to these as the ``human expert ratings'' and use them as a reference to measure the accuracy of LLM ratings.

\subsection{Source Political Bias Score}

To quantify partisan bias of the news sources in the US political context, we leverage the scores generated by \citet{robertson2018auditingpartisan}.
The authors first construct a panel of over half a million Twitter accounts matched with US voter registration records and then measure the partisan bias of the audience of each source.
The results are scores between $-1$ and $+1$, where a score of $-1$ means that a source is shared exclusively by Democrats, and a score of $+1$ means that a source is shared exclusively by Republicans.
The authors show that their scores correlate highly with other ratings, such as the ones produced by \citet{bakshy2015exposure}, \citet{budak2016fair}, and \url{allsides.com}.
Merging the list from \citet{robertson2018auditingpartisan} with the human expert ratings of credibility, we obtain 2,683 sources with political leanings to analyze the political biases of LLMs.

\section{Results}

\subsection{LLM Response Analysis}

As described in the Methods section, we first run the 7,523 news sources through all nine LLMs using the same prompt in the default configuration (no partisan roles).
In most cases, the LLMs respond with the required content in the specified format.
When errors occur, we rerun the queries until we obtain the desired outputs.
The code and data are available at our GitHub repository\footnote{\url{github.com/osome-iu/llm_domain_rating}}.

Take \url{reuters.com} as an example. GPT-4o provides the following response:
``\textbf{Rating}: 1.0; \textbf{Explanation}: Reuters is a well-established international news organization known for its accurate and unbiased reporting. It is widely respected in the journalism community and has a long history of providing reliable news.''
All other models give Reuters credibility scores over 0.9 with similar explanations (complete responses available in the Appendix).
These responses indicate that LLMs recognize news outlets from their websites, possess information about them, and can provide credibility ratings.

\begin{figure}
    \centering
    \includegraphics[width=\columnwidth]{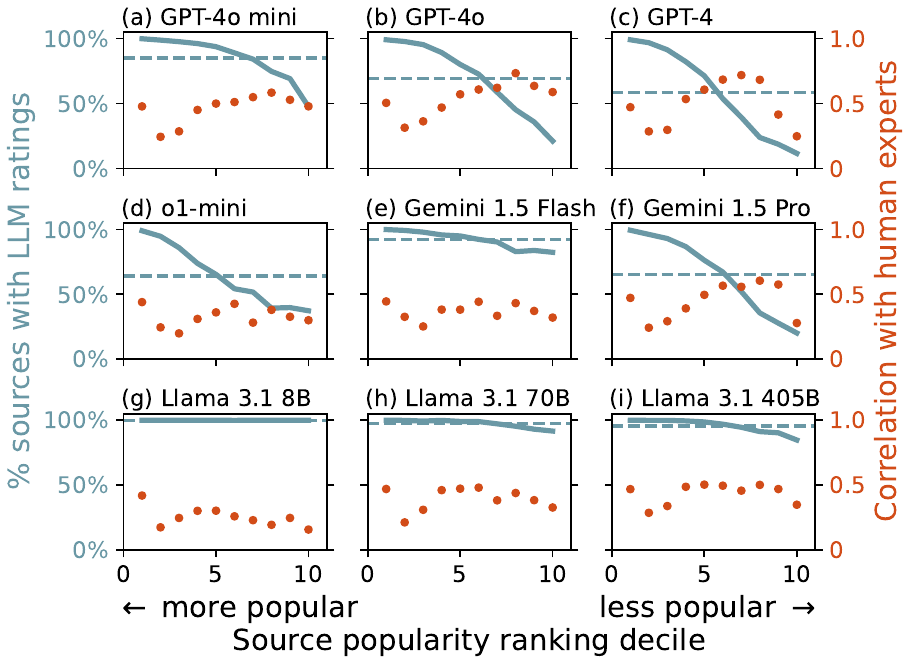}
    \caption{
    Relationship between source popularity and the responses of LLMs.
    The left axes and the lines show the percentages of sources for which each LLM provides ratings.
    The dashed lines indicate the overall percentages, whereas the solid lines illustrate the results for different source ranking deciles.
    The right axes and the dots represent the Spearman correlation coefficients between LLM ratings and human expert ratings in different source ranking deciles.
    Sources in larger ranking deciles are less popular.
    }
    \Description{Relationship between source popularity and the responses of LLMs.}
    \label{fig:byrank}
\end{figure}

When the LLMs lack sufficient information about a given source, they respond with a rating of $-1$, as instructed.
In Figure~\ref{fig:byrank}, we illustrate the percentage of sources for which each LLM provides ratings (dashed lines).
We find that within each family, larger models are more likely to refuse to rate sources due to insufficient information.
We hypothesize that LLMs tend to lack knowledge about unpopular news sources.
To confirm this, we split the sources into 10 deciles based on the popularity ranking and calculate the percentage of sources with LLM ratings within each decile.
The solid lines in Figure~\ref{fig:byrank} support our hypothesis for all models.

\begin{figure}
    \centering
    \includegraphics[width=0.8\columnwidth]{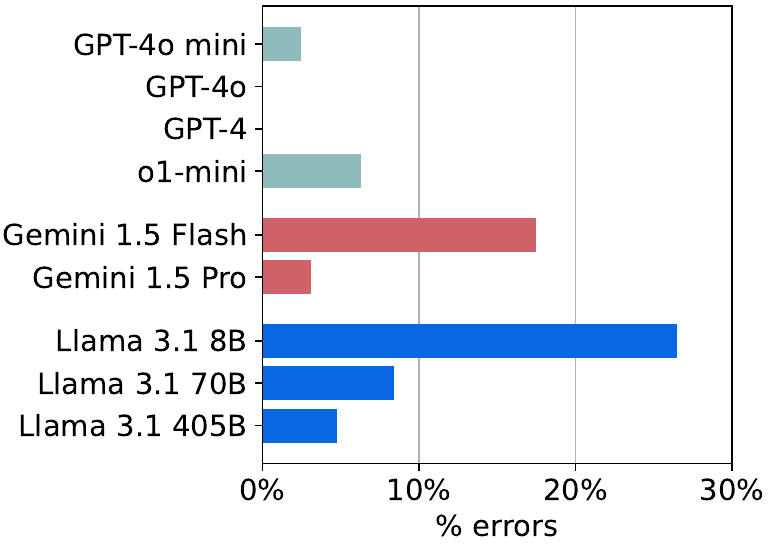}
    \caption{
    Percentage of errors among 200 manually annotated cases for each LLM.
    }
    \Description{Percentage of errors among 200 manually annotated cases for each LLM.}
    \label{fig:hallucination_rate}
\end{figure}

Figure~\ref{fig:byrank} also shows that Llama models can provide ratings for more sources compared to OpenAI and Gemini models.
However, LLMs are known to suffer from hallucinations, where they generate baseless responses to user requests~\cite{ji2023survey}.
To gain deeper insights into the ratings generated by these LLMs, we randomly select 200 sources and manually analyze the responses from all nine models.
Through this process, we identify two common types of unambiguous errors.
First, the LLMs can mistakenly associate a website with the wrong organization.
For example, five out of nine LLMs identify \url{aldf.com}, the official website of the ``American Lyme Disease Foundation,'' as belonging to the ``Animal Legal Defense Fund'' (\url{aldf.org}).
The complete responses of the LLMs can be found in the Appendix.
Second, an LLM can fail to identify a website and still generate a rating for it.

In Figure~\ref{fig:hallucination_rate}, we illustrate the percentage of unambiguous errors among the annotated cases for each LLM.
The results show that smaller models tend to make more errors within each family.
In the OpenAI family, o1-mini exhibits the highest rate of errors among all models.
However, as mentioned in the Methods section, its number of parameters is not comparable to the other models due to its special design.
Across different providers, Llama and Gemini models exhibit higher rates of errors in general.
It is important to note that even when the models correctly recognize the sources, they may still provide inaccurate ratings due to other issues.
Next, we assess the accuracy of the LLM ratings by comparing them with those given by human experts.

\subsection{Rating Accuracy}

\begin{figure}
    \centering
    \includegraphics[width=\columnwidth]{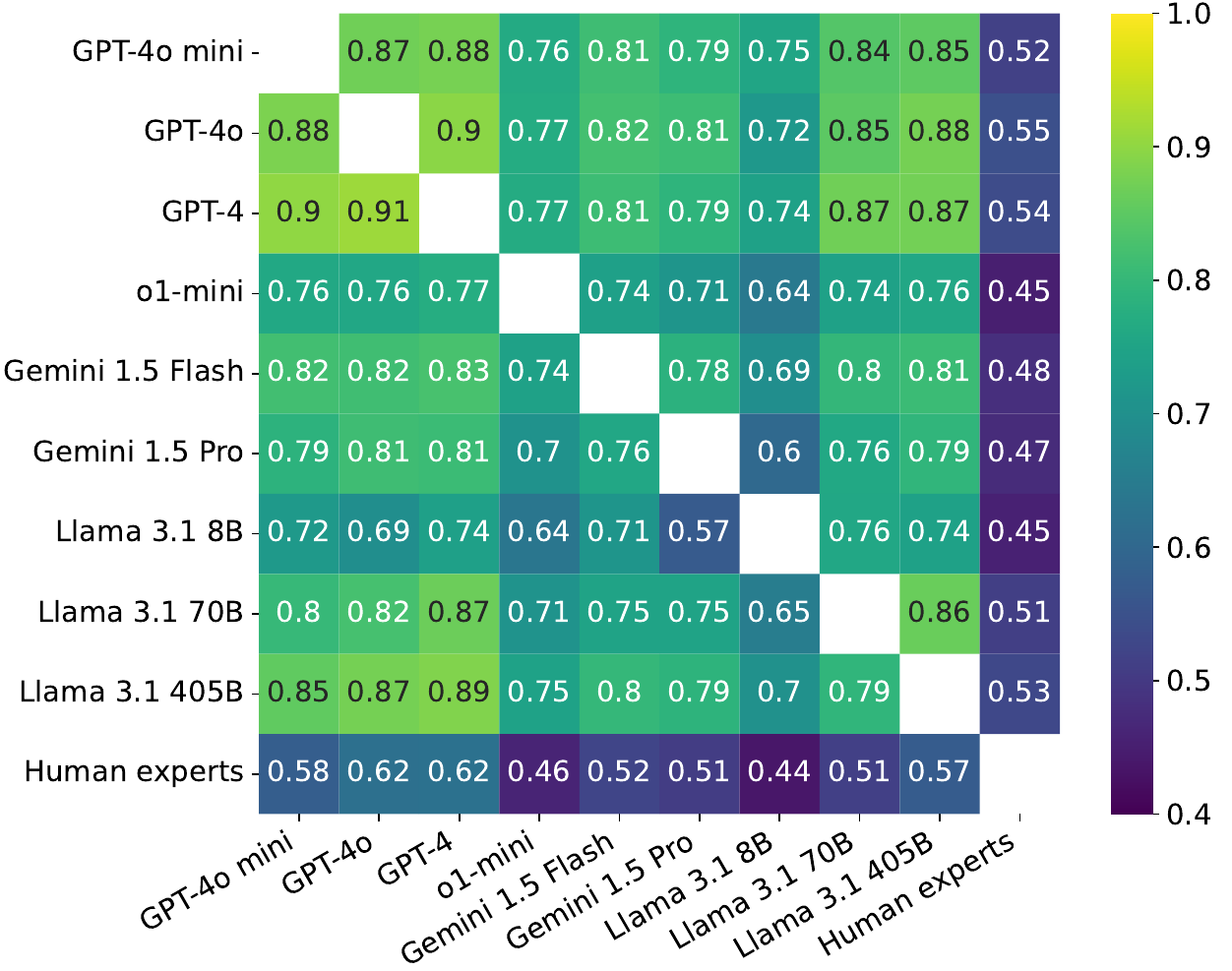}
    \caption{
    Heatmap of source credibility rating correlation (Spearman's $\rho$) among different LLMs and human experts.
    Results in the upper right triangle of the heatmap are based on 3,077 (40.9\%) sources rated by all LLMs.
    Results in the lower left triangle are based on the sources rated by both raters in comparison.
    }
    \Description{Heatmap of source credibility rating correlation (Spearman) among different LLMs and human experts.}
    \label{fig:correlations}
\end{figure}

Let us measure how well the LLM ratings correlate with each other and align with those from human experts.
For each pair of raters (LLMs or human experts), we calculate the Spearman rank correlation coefficient $\rho$ between their ratings.
Other metrics of performance, such as accuracy and F1 scores, are not used because they typically require a threshold to determine the positive/negative cases, which can be arbitrary in the present context.
And because the distributions of the ratings are not normal (see visualizations in the Appendix), we use the Spearman rank correlation coefficient instead of the Pearson correlation coefficient.

Given that some LLMs only provide ratings for a subset of the sources, we use two different approaches to calculate the correlation coefficients.
First, we include only the 3,077 (40.9\%) sources that are rated by all LLMs.
Second, for each pair of raters, we focus on their intersection.
The results from both approaches are shown in Figure~\ref{fig:correlations} and demonstrate similar patterns.
Therefore, our discussion below will focus on the results from the first approach.

All correlation coefficients in Figure~\ref{fig:correlations} are positive and statistically significant with $p<0.001$.
We find a high agreement level among LLMs, with an average $\rho=0.79$, despite their different providers.
Conversely, they only moderately correlate with human experts, with an average $\rho=0.50$ and small variation across models.

To provide context for our findings, we compare them with the recent work of \citet{williams2024bridgingsocialmediasearch}.
Their study combines social media data and web graph analysis to identify unreliable domains, achieving slightly better performance (accuracy=0.819) than GPT-3.5 (accuracy=0.782) on the same set of domains used in our study.
While a direct comparison is not possible due to the unavailability of their code and data, we test GPT-3.5 (gpt-3.5-turbo-0125) under our experimental conditions.
The model yields a Spearman correlation coefficient of 0.50 with human expert ratings, which is the same as the average correlation coefficient across all LLMs in our study.

We also test the impact of website popularity on the accuracy of LLM ratings by splitting the sources into 10 popularity ranking deciles and calculating rating correlations between each LLM and human experts.
The results, shown as dots in Figure~\ref{fig:byrank}, do not indicate a clear association between LLMs accuracy and source popularity.

\subsection{Political Biases}

To probe political biases in LLM ratings, we focus on the 2,683 sources with partisan leanings in the US context.
We query the LLMs with the same prompts except for assigning them different partisan roles.
Specifically, for each source, we instruct the models to rate its credibility from the viewpoints of Democrats, Independents, and Republicans, respectively.

\begin{figure}[t]
    \centering
    \includegraphics[width=\columnwidth]{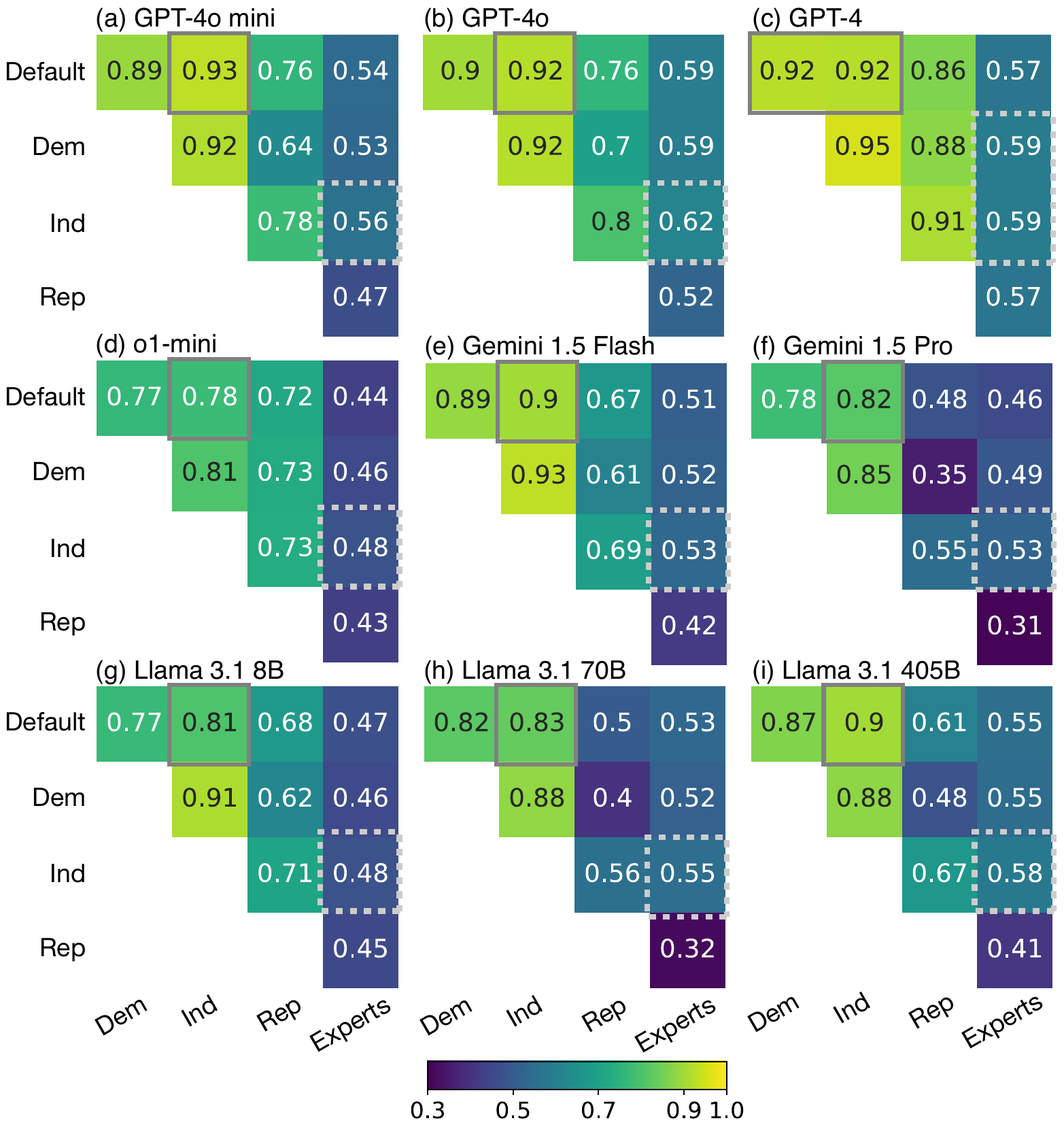}
    \caption{
    Heatmaps of Spearman correlation coefficients among the ratings generated by LLMs with different partisan roles.
    The highest correlation coefficients between the default LLM configuration and different partisan roles are highlighted by squares with solid edges.
    The highest correlation coefficients between human experts and different partisan roles are highlighted by squares with dashed edges.
    }
    \Description{Heatmaps of Spearman correlation coefficients among the ratings generated by LLMs with different partisan roles.}
    \label{fig:identity}
\end{figure}

In Figure~\ref{fig:identity}, we show the correlations between the ratings generated by LLMs with different partisan roles and those from human experts.
Ratings from LLMs in the default configuration (no partisan roles) are closest to those from the Independent role, followed by Democratic roles.
The correlations with the LLMs identified as Republicans are the lowest.
When compared with human experts, the LLMs identified as Independents show higher correlations than all other cases, including the default ones.
These patterns are consistent across all models.

To quantify the partisan biases of LLMs with different partisan roles, we calculate the \textit{LLM rating bias score} for each source, defined as the difference between the LLM rating and the human expert rating.
This metric accounts for the fact that left-leaning sources in our dataset tend to have higher human expert ratings.
A positive/negative bias score means the LLM considers the source more/less credible than expected.

\begin{figure}[t]
    \centering
    \includegraphics[width=0.9\columnwidth]{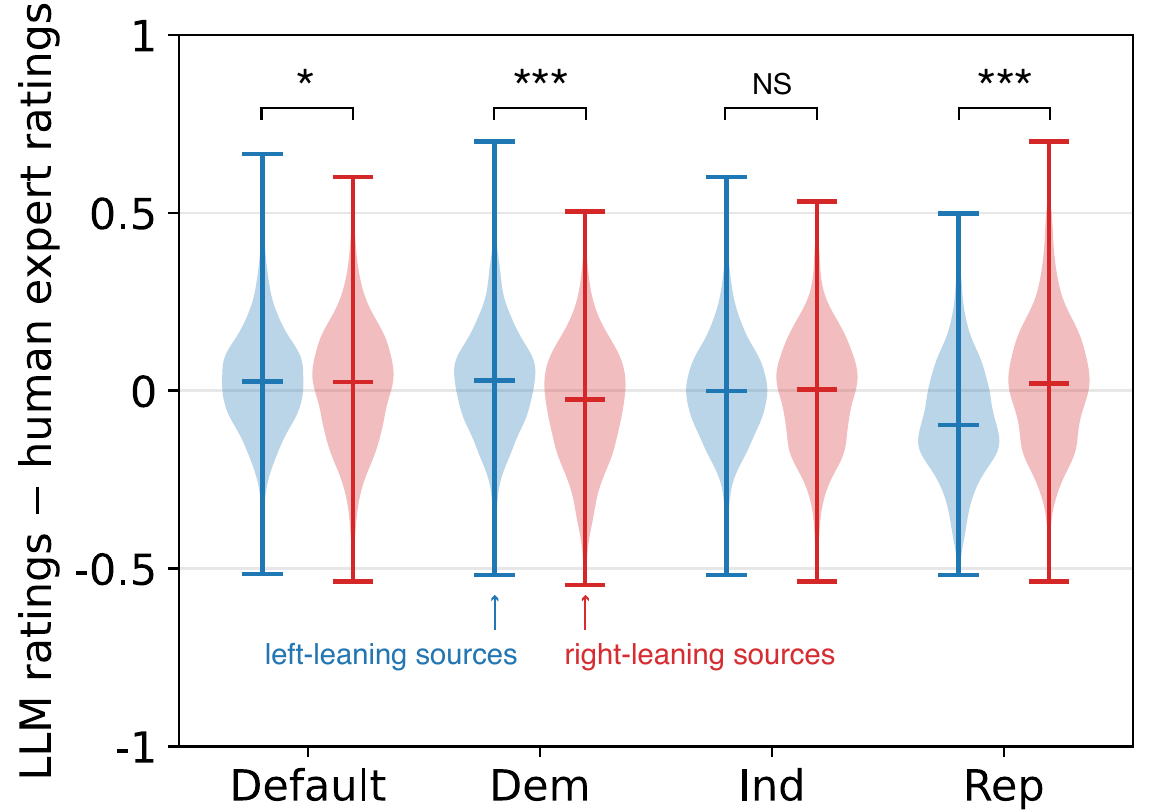}
    \caption{
    Distributions of LLM rating bias scores of GPT-4o mini with different partisan roles.
    The blue and red violins represent the results for left- and right-leaning sources, respectively.
    Significance of t-tests is indicated by ***: $p<0.001$, *:$p<0.05$, NS: not significant.
    }
    \Description{Distributions of LLM rating bias scores of GPT-4o mini with different partisan roles.}
    \label{fig:llm_rating_bias_score}
\end{figure}

Based on the scores provided by \citet{robertson2018auditingpartisan}, we classify the 2,683 sources as left- or right-leaning using zero as the threshold and compare their LLM rating bias scores.
Note that \citet{chen2021neutral} suggest using 0.058 as the threshold, which leads to qualitatively similar results in our experiments according to robustness checks.

In Figure~\ref{fig:llm_rating_bias_score}, we show the distributions of LLM rating bias scores for GPT-4o mini with different partisan roles.
We find that the default configuration and the Democratic role exhibit liberal bias and are more likely to assign higher-than-expected credibility scores to left-leaning sources.
The Republican role, on the other hand, favors right-leaning sources.
The Independent role shows no significant differences in LLM rating bias scores between left- and right-leaning sources.

\begin{figure}[t]
    \centering
    \includegraphics[width=0.85\columnwidth]{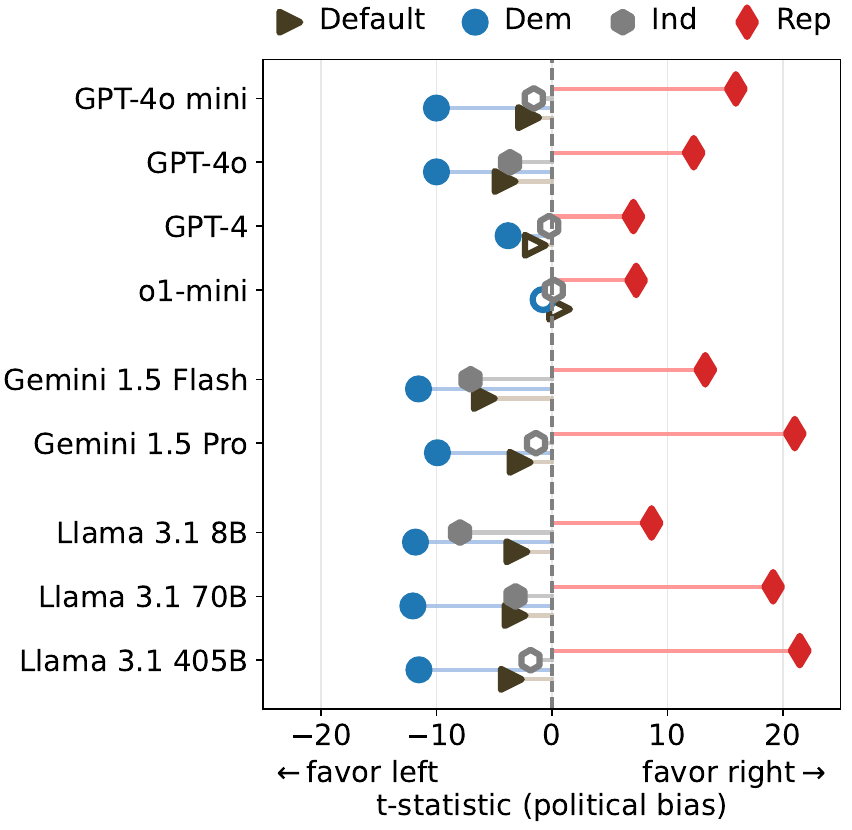}
    \caption{
    Political biases of different LLM-role configurations.
    The political biases are measured by t-statistics between the distributions of LLM rating bias scores on the left- and right-leaning sources.
    A negative/positive t-statistic means the LLM-role configuration favors left-/right-leaning news outlets.
    The solid symbols indicate statistically significant differences ($p<0.05$).
    }
    \Description{Political biases of different LLM-role configurations.}
    \label{fig:political_bias}
\end{figure}

We replicate the analysis for all LLMs and show the results in Figure~\ref{fig:political_bias}.
Due to space constraints, we only report the t-statistics from the tests comparing the distributions of LLM rating bias scores for left- and right-leaning news sources.
We find that assigning a Democratic role leads to a liberal bias for all models except for o1-mini, whereas assigning a Republican role leads to a conservative bias for all models.
The Independent role shows no significant political biases for five models but a liberal bias for the other four models, although less pronounced than the Democratic role.
Under the default configuration, where no explicit partisan roles are assigned to the LLMs, all models except for GPT-4 and o1-mini show a statistically significant liberal bias.

\subsection{Political Bias and Rating Accuracy}

\begin{figure}[t]
    \centering
    \includegraphics[width=0.85\columnwidth]{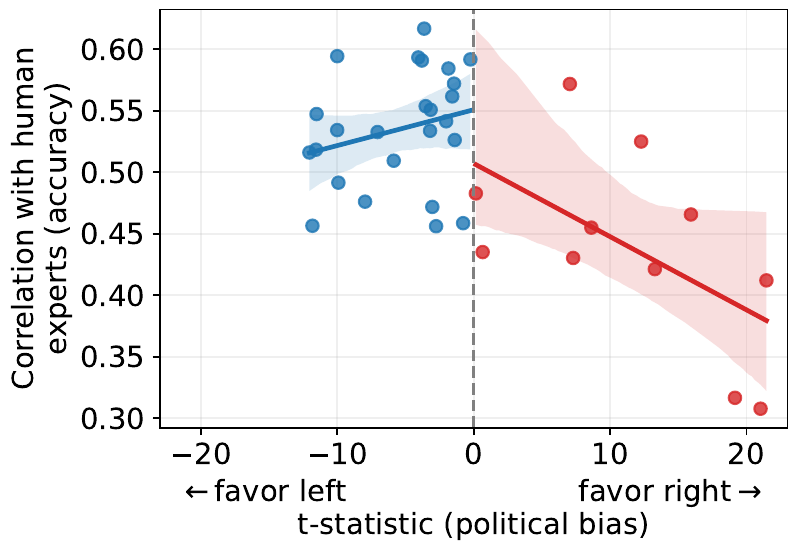}
    \caption{
    Political bias vs. rating accuracy for all LLM-role configurations.
    We use the t-statistics from the tests comparing the distributions of LLM rating bias scores on left- and right-leaning sources to quantify the political bias and the correlation with human experts for accuracy.
    LLM-role configurations with left or right biases are separated and the lines represent linear regressions on the two groups.
    For the left-biased data points, we have the Spearman correlation coefficient between the political bias and the rating accuracy of $\rho=0.18$ ($p=0.38$).
    For the right-biased data points, we have $\rho=-0.67$ ($p=0.02$).
    }
    \Description{Political bias vs. rating accuracy for all LLM-role configurations.}
    \label{fig:bias_vs_accuracy}
\end{figure}

The results in Figures~\ref{fig:identity} and \ref{fig:political_bias} suggest a negative correlation between the political biases and the accuracy of LLM-role configurations.
To confirm this, we present a scatter plot of all model-role configurations in Figure~\ref{fig:bias_vs_accuracy}.
Regardless of the direction, we find that a stronger political bias correlates with lower rating accuracy, although such correlation is only statistically significant for the cases where the LLM-role configuration is right-biased.
This finding suggests that the misalignment between LLMs and human experts is partially due to the political biases embedded in LLMs and that reducing these biases could enhance rating accuracy.

\citet{pennycook2019fighting} ran an experiment in which participants produced news source ratings that correlated with their different ideologies.
However, the combined ratings from participants on both sides correlate strongly with expert ratings.
Given that LLMs exhibit consistent biases congruent with assigned roles, we test whether combining ratings from LLMs with different viewpoints could reduce overall biases and improve accuracy.

We consider two aggregate ratings for each model: (1)~D+R, the average ratings of the Democratic and Republican roles, and (2)~D+I+R, the average ratings of the Democratic, Independent, and Republican roles.
Although different weights can be assigned to these partisan roles, we choose equal weights to reflect the composition of American voters with varying partisanship.\footnote{pewresearch.org/politics/2024/04/09/the-partisanship-and-ideology-of-american-voters}

\begin{figure}
    \centering
    \includegraphics[width=\columnwidth]{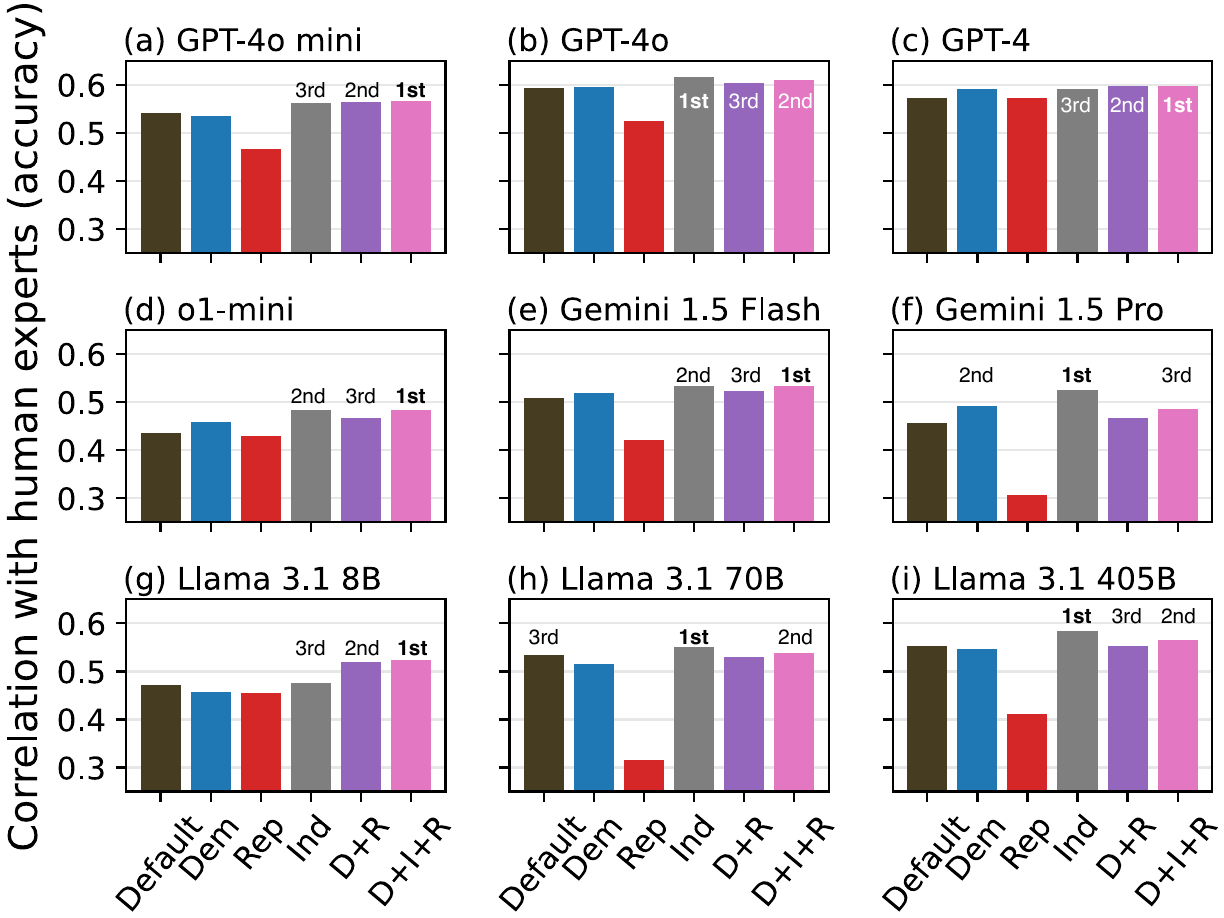}
    \caption{
    Accuracy of different LLM ratings.
    We show the Spearman correlation coefficients between different LLM ratings and those from human experts.
    For each model, we report the results for the default configuration, three partisan roles (Democrat, Independent, and Republican), as well as two aggregate ratings, D+R (average of Democratic and Republican roles) and D+I+R (average of all three political roles).
    The top three configurations are annotated.
    }
    \Description{Accuracy of different LLM ratings.}
    \label{fig:crowdsource}
\end{figure}

In Figure~\ref{fig:crowdsource}, we compare the accuracy of D+R and D+I+R with the results for different partisan roles.
In most cases, D+R and D+I+R ratings are more accurate than either Democratic or Republican ratings, suggesting that blending different perspectives leads to better judgment of source credibility.
Across all configurations, the Independent role is the winner for four models, while D+I+R leads in the other four cases.
However, the differences between the Independent and D+I+R ratings are marginal.
We, therefore, conjecture that aggregate ratings from partisan roles do not lead to significant improvement over the Independent role, which already exhibits relatively low political bias.

\section{Discussion}

\subsection{Summary of Results}

In this paper, we systematically audit nine widely used LLMs from three leading providers to test their ability to discern credible information sources from low-credibility ones.
For most news sources tested in our experiments, the LLMs could provide ratings as instructed.
However, larger models tend to refuse to rate less popular sources due to insufficient knowledge, while smaller models tend to make more errors.
Comparing the LLM ratings, these models exhibit a high level of agreement despite being trained by different providers.
Conversely, the models only moderately correlate with human experts.

It remains unclear how the LLMs acquire this capacity.
Based on the explanations provided by the models alongside the ratings, we hypothesize that the models summarize descriptions of the given news sources from their training data and generate ratings accordingly.
For instance, when encountering high-credibility sources, the LLMs often note that these are well-established and reputable news websites.
This could explain the high correlation among the LLMs, as they likely have common training data~\cite{liu2024datasetslargelanguagemodels}.

Focusing on news sources with clear political leanings in the US context, we find that assigning partisan roles to LLMs steers their ratings to favor sources with congruent leanings in most cases.
This result indicates that LLMs can reflect the viewpoints of humans with different political ideologies~\cite{argyle2023out,simmons2023moralmimicrylargelanguage}.
In their default configurations, most LLMs exhibit a liberal bias: they are more likely to assign higher-than-expected credibility ratings for left-leaning sources compared to right-leaning ones.
LLMs assigned with an Independent role are the least biased but still show a weak liberal bias.
These findings align with previous studies indicating that many LLMs have left-leaning tendencies~\cite{santurkar2023whose}.

We also explore approaches to reduce political biases and increase the accuracy of the LLM ratings.
We find that explicitly assigning an Independent role to an LLM or mixing the ratings from different roles could help achieve both objectives simultaneously.
However, even unbiased ratings fail to align perfectly with human judgment.

\subsection{Limitations}

Our auditing has some limitations.
First, since the evaluation relies on existing human ratings, any biases in these ratings may propagate into our findings.
For example, our binary classification of news sources as left- or right-leaning based on a zero threshold applied to the scores by \citet{robertson2018auditingpartisan} simplifies our bias analysis, but the alternative 5-class approach used by organizations like MBFC might provide more nuanced results.

In our experiments, we only provide the LLMs with the domains of the news sources, which is useful for assessing the completeness and accuracy of the internal knowledge.
However, in real-world scenarios, the models are likely to have access to additional information about the sources, such as the metadata and content of the pages, which could potentially improve their performance~\cite{schlichtkrull2024generatingmediabackgroundchecks}.

There are different approaches to prompt the LLMs, which might yield different outcomes.
For instance, one could employ a binary classification approach or ask LLMs to rank two sources at a time~\cite{qin2024largelanguagemodelseffective}.
When assigning a partisan role to the LLMs, one could also provide them with detailed demographics and background information about a persona.
Additionally, different prompt engineering techniques could be deployed to boost the performance of the LLMs.
We were unable to test all of these approaches.

Our findings might not be generalizable to all LLMs and contexts.
Despite our efforts to test as many representative models as possible, we are unable to cover all the other LLMs from different providers on the market.
Given the rapid development in the field, new models with different behaviors will emerge soon.
Finally, our experiments mainly focus on the US context, which might not be representative of other countries and non-English news.

\subsection{Implications}

Despite these limitations, we believe our findings provide valuable insights into the current state-of-the-art LLMs, considering that the general patterns are consistent regardless of the model size and provider.
Given that these LLMs have been deployed in widely used systems, the findings have important implications.

Our experiments show that LLMs have the potential to serve as an affordable and accessible reference for source credibility ratings.
Such ratings are vital for researchers to study the dissemination and prevalence of misinformation online~\cite{lazer2018science,lin2023high}.
With the ability to provide contextual information and actively answer user questions, LLMs might also be adopted by the general public as scalable media literacy tools to investigate the credibility of news sources and perform fact-checking~\cite{pennycook2019fighting,hoes2023chatgpt}.
However, our results indicate that the accuracy of LLMs is not perfect, calling for further analysis and comparison with other misinformation intervention methods~\cite{kozyreva2024toolbox}.
Additionally, LLMs may be vulnerable to manipulation.
For instance, bad actors could pollute LLM training data by creating fake news source review websites.

Our findings also highlight the challenges of using LLMs as information curators.
Although not confirmed by the providers, leaked system prompts of popular LLMs suggest that they are configured to retrieve information according to the quality and credibility of sources when searching the Internet, along with other criteria such as the relevance to user questions and the freshness of the information.\footnote{github.com/jujumilk3/leaked-system-prompts}
However, it is unclear how the LLMs understand and implement these criteria in production.

A famous failure story came from Google in 2024 when its AI overview feature produced bizarre answers such as recommending that people eat rocks.\footnote{bbc.com/news/articles/cd11gzejgz4o}
According to Google, these odd responses originated from low-quality information sources,\footnote{blog.google/products/search/ai-overviews-update-may-2024} yet their AI models appeared to overlook it.
Although these errors may have a limited negative impact, they highlight the risks of using LLMs for information curation in high-stakes contexts such as public health and elections.

Our findings provide valuable insights about such risks.
We show that LLMs may lack information about many information sources, especially unpopular ones.
However, it is inevitable for LLMs to encounter unfamiliar sources when curating information for users, especially when they search for misinformation topics~\cite{boyd2018data,aslett2024online}.
Under these circumstances, forcing the models to produce summaries or responses could inadvertently amplify lesser-known low-credibility sources.
Even when the models know the information sources, including the highly popular ones, their assessment of credibility still deviates from human judgment.
This inaccuracy might suppress credible sources and amplify low-credibility ones.
When dealing with political information, the inherent bias of the LLMs in the default configuration might lead to distorted outcomes.
The finding that assigning partisan roles to LLMs could induce corresponding biases suggests that LLMs tailored to user preferences and views might exacerbate echo chambers and polarization.

\subsection{Future Studies}

For future research, a critical challenge lies in mitigating biases and enhancing the accuracy of LLMs in assessing source credibility.
While one might expect that models demonstrating superior performance on standard benchmarks would naturally excel at this specific task, our findings suggest otherwise.
Our comparative analysis of models within the same family, such as the GPT series, reveals that increased model size and parameter count do not consistently translate to improved accuracy or reduced biases.
Although OpenAI's o1 model series represents an innovative approach by training models specifically for complex reasoning tasks, our evaluation of o1-mini indicates that this enhanced reasoning capability has not effectively transferred to the present task.
These findings suggest that more capable LLMs in the future might not necessarily perform better on the source credibility assessment task.

In our attempt to address this challenge, we leverage the prompt-based approach by assigning partisan roles to the LLMs and aggregating ratings from multiple roles.
This approach is simple and cheap, and can be applied to both proprietary and open-source models, but the improvement is limited.
Future models might consider explicitly integrating source credibility information into the models either through fine-tuning or including the data in the pre-training phase~\cite{ouyang2022training}.
When embedding LLMs into larger systems such as search engines, one may also consider providing the models with the credibility ratings of the sources together with the content and query for reasoning.

Another key challenge is to enhance LLMs' ability to evaluate previously unseen information sources.
Our findings demonstrate that LLMs struggle to assess the credibility of less popular sources, significantly limiting their effectiveness as information curators.
While human expert evaluation could provide accurate ratings, this approach is neither sustainable nor scalable, given the vast number of sources and limited expert resources.
Future research should explore automated frameworks that leverage LLMs to assess source credibility~\cite{schlichtkrull2024generatingmediabackgroundchecks}.
Such frameworks could autonomously gather information on the sources from the Internet, leveraging advanced reasoning capabilities to process such information, and generate reliable credibility ratings.

Other important directions include exploring how LLMs handle information from different sources in more realistic scenarios, how other information selection criteria, such as relevance and freshness~\cite{sun2023chatgptgoodsearchinvestigating}, affect the outcomes, and how to mitigate emerging manipulation techniques~\cite{pfrommer2024rankingmanipulationconversationalsearch}.
At the same time, more studies on how humans interact with AI-augmented systems~\cite{sharma2024generative,deverna2024factchecking} and how information processed by AI influences humans are needed~\cite{jakesch2023co,simchon2024persuasive}.

\bibliographystyle{ACM-Reference-Format}
\bibliography{refs}


\begin{thebibliography}{52}


\ifx \showCODEN    \undefined \def \showCODEN     #1{\unskip}     \fi
\ifx \showDOI      \undefined \def \showDOI       #1{#1}\fi
\ifx \showISBNx    \undefined \def \showISBNx     #1{\unskip}     \fi
\ifx \showISBNxiii \undefined \def \showISBNxiii  #1{\unskip}     \fi
\ifx \showISSN     \undefined \def \showISSN      #1{\unskip}     \fi
\ifx \showLCCN     \undefined \def \showLCCN      #1{\unskip}     \fi
\ifx \shownote     \undefined \def \shownote      #1{#1}          \fi
\ifx \showarticletitle \undefined \def \showarticletitle #1{#1}   \fi
\ifx \showURL      \undefined \def \showURL       {\relax}        \fi
\providecommand\bibfield[2]{#2}
\providecommand\bibinfo[2]{#2}
\providecommand\natexlab[1]{#1}
\providecommand\showeprint[2][]{arXiv:#2}

\bibitem[Abid et~al\mbox{.}(2021)]%
        {abid2021persistent}
\bibfield{author}{\bibinfo{person}{Abubakar Abid}, \bibinfo{person}{Maheen Farooqi}, {and} \bibinfo{person}{James Zou}.} \bibinfo{year}{2021}\natexlab{}.
\newblock \showarticletitle{Persistent Anti-Muslim Bias in Large Language Models}. In \bibinfo{booktitle}{\emph{Proceedings of the 2021 AAAI/ACM Conference on AI, Ethics, and Society}} (Virtual Event, USA) \emph{(\bibinfo{series}{AIES '21})}. \bibinfo{publisher}{Association for Computing Machinery}, \bibinfo{address}{New York, NY, USA}, \bibinfo{pages}{298–306}.
\newblock
\showISBNx{9781450384735}


\bibitem[Argyle et~al\mbox{.}(2023)]%
        {argyle2023out}
\bibfield{author}{\bibinfo{person}{Lisa~P. Argyle}, \bibinfo{person}{Ethan~C. Busby}, \bibinfo{person}{Nancy Fulda}, \bibinfo{person}{Joshua~R. Gubler}, \bibinfo{person}{Christopher Rytting}, {and} \bibinfo{person}{David Wingate}.} \bibinfo{year}{2023}\natexlab{}.
\newblock \showarticletitle{Out of One, Many: Using Language Models to Simulate Human Samples}.
\newblock \bibinfo{journal}{\emph{Political Analysis}} \bibinfo{volume}{31}, \bibinfo{number}{3} (\bibinfo{year}{2023}), \bibinfo{pages}{1–15}.
\newblock


\bibitem[Aslett et~al\mbox{.}(2024)]%
        {aslett2024online}
\bibfield{author}{\bibinfo{person}{Kevin Aslett}, \bibinfo{person}{Zeve Sanderson}, \bibinfo{person}{William Godel}, \bibinfo{person}{Nathaniel Persily}, \bibinfo{person}{Jonathan Nagler}, {and} \bibinfo{person}{Joshua~A Tucker}.} \bibinfo{year}{2024}\natexlab{}.
\newblock \showarticletitle{Online searches to evaluate misinformation can increase its perceived veracity}.
\newblock \bibinfo{journal}{\emph{Nature}} \bibinfo{volume}{625}, \bibinfo{number}{7995} (\bibinfo{year}{2024}), \bibinfo{pages}{548--556}.
\newblock


\bibitem[Bai et~al\mbox{.}(2024)]%
        {bai2024measuringimplicitbiasexplicitly}
\bibfield{author}{\bibinfo{person}{Xuechunzi Bai}, \bibinfo{person}{Angelina Wang}, \bibinfo{person}{Ilia Sucholutsky}, {and} \bibinfo{person}{Thomas~L. Griffiths}.} \bibinfo{year}{2024}\natexlab{}.
\newblock \bibinfo{title}{Measuring Implicit Bias in Explicitly Unbiased Large Language Models}.
\newblock
\newblock
\showeprint[arxiv]{2402.04105}~[cs.CY]
\urldef\tempurl%
\url{https://arxiv.org/abs/2402.04105}
\showURL{%
\tempurl}


\bibitem[Bakshy et~al\mbox{.}(2015)]%
        {bakshy2015exposure}
\bibfield{author}{\bibinfo{person}{Eytan Bakshy}, \bibinfo{person}{Solomon Messing}, {and} \bibinfo{person}{Lada~A Adamic}.} \bibinfo{year}{2015}\natexlab{}.
\newblock \showarticletitle{Exposure to ideologically diverse news and opinion on Facebook}.
\newblock \bibinfo{journal}{\emph{Science}} \bibinfo{volume}{348}, \bibinfo{number}{6239} (\bibinfo{year}{2015}), \bibinfo{pages}{1130--1132}.
\newblock


\bibitem[Boyd and Golebiewski(2018)]%
        {boyd2018data}
\bibfield{author}{\bibinfo{person}{Danah Boyd} {and} \bibinfo{person}{Michael Golebiewski}.} \bibinfo{year}{2018}\natexlab{}.
\newblock \bibinfo{title}{Data voids: Where missing data can easily be exploited}.
\newblock
\newblock


\bibitem[Budak et~al\mbox{.}(2016)]%
        {budak2016fair}
\bibfield{author}{\bibinfo{person}{Ceren Budak}, \bibinfo{person}{Sharad Goel}, {and} \bibinfo{person}{Justin~M Rao}.} \bibinfo{year}{2016}\natexlab{}.
\newblock \showarticletitle{Fair and balanced? Quantifying media bias through crowdsourced content analysis}.
\newblock \bibinfo{journal}{\emph{Public Opinion Quarterly}} \bibinfo{volume}{80}, \bibinfo{number}{S1} (\bibinfo{year}{2016}), \bibinfo{pages}{250--271}.
\newblock


\bibitem[Carragher et~al\mbox{.}(2024)]%
        {carragher2024detection}
\bibfield{author}{\bibinfo{person}{Peter Carragher}, \bibinfo{person}{Evan~M. Williams}, {and} \bibinfo{person}{Kathleen~M. Carley}.} \bibinfo{year}{2024}\natexlab{}.
\newblock \showarticletitle{Detection and Discovery of Misinformation Sources Using Attributed Webgraphs}.
\newblock \bibinfo{journal}{\emph{Proceedings of the International AAAI Conference on Web and Social Media}} \bibinfo{volume}{18}, \bibinfo{number}{1} (\bibinfo{year}{2024}), \bibinfo{pages}{214--226}.
\newblock


\bibitem[Chen et~al\mbox{.}(2021)]%
        {chen2021neutral}
\bibfield{author}{\bibinfo{person}{Wen Chen}, \bibinfo{person}{Diogo Pacheco}, \bibinfo{person}{Kai-Cheng Yang}, {and} \bibinfo{person}{Filippo Menczer}.} \bibinfo{year}{2021}\natexlab{}.
\newblock \showarticletitle{Neutral bots probe political bias on social media}.
\newblock \bibinfo{journal}{\emph{Nature communications}} \bibinfo{volume}{12}, \bibinfo{number}{1} (\bibinfo{year}{2021}), \bibinfo{pages}{5580}.
\newblock


\bibitem[Cheng et~al\mbox{.}(2023)]%
        {cheng2023markedpersonasusingnatural}
\bibfield{author}{\bibinfo{person}{Myra Cheng}, \bibinfo{person}{Esin Durmus}, {and} \bibinfo{person}{Dan Jurafsky}.} \bibinfo{year}{2023}\natexlab{}.
\newblock \showarticletitle{Marked Personas: Using Natural Language Prompts to Measure Stereotypes in Language Models}. In \bibinfo{booktitle}{\emph{Proceedings of the 61st Annual Meeting of the Association for Computational Linguistics}}. \bibinfo{publisher}{Association for Computational Linguistics}, \bibinfo{address}{Toronto, Canada}, \bibinfo{pages}{1504--1532}.
\newblock


\bibitem[Cinelli et~al\mbox{.}(2021)]%
        {cinelli2021echo}
\bibfield{author}{\bibinfo{person}{Matteo Cinelli}, \bibinfo{person}{Gianmarco De~Francisci Morales}, \bibinfo{person}{Alessandro Galeazzi}, \bibinfo{person}{Walter Quattrociocchi}, {and} \bibinfo{person}{Michele Starnini}.} \bibinfo{year}{2021}\natexlab{}.
\newblock \showarticletitle{The echo chamber effect on social media}.
\newblock \bibinfo{journal}{\emph{Proceedings of the National Academy of Sciences}} \bibinfo{volume}{118}, \bibinfo{number}{9} (\bibinfo{year}{2021}), \bibinfo{pages}{e2023301118}.
\newblock


\bibitem[Conover et~al\mbox{.}(2021)]%
        {conover2021political}
\bibfield{author}{\bibinfo{person}{Michael Conover}, \bibinfo{person}{Jacob Ratkiewicz}, \bibinfo{person}{Matthew Francisco}, \bibinfo{person}{Bruno Goncalves}, \bibinfo{person}{Filippo Menczer}, {and} \bibinfo{person}{Alessandro Flammini}.} \bibinfo{year}{2021}\natexlab{}.
\newblock \showarticletitle{Political Polarization on Twitter}.
\newblock \bibinfo{journal}{\emph{Proceedings of the International AAAI Conference on Web and Social Media}} \bibinfo{volume}{5}, \bibinfo{number}{1} (\bibinfo{year}{2021}), \bibinfo{pages}{89--96}.
\newblock


\bibitem[DeVerna et~al\mbox{.}(2024)]%
        {deverna2024factchecking}
\bibfield{author}{\bibinfo{person}{Matthew~R. DeVerna}, \bibinfo{person}{Harry Yan}, \bibinfo{person}{Kai-Cheng Yang}, {and} \bibinfo{person}{Filippo Menczer}.} \bibinfo{year}{2024}\natexlab{}.
\newblock \showarticletitle{Fact-checking information from large language models can decrease headline discernment}.
\newblock \bibinfo{journal}{\emph{Proceedings of the National Academy of Sciences}} \bibinfo{volume}{121}, \bibinfo{number}{50} (\bibinfo{year}{2024}), \bibinfo{pages}{e2322823121}.
\newblock


\bibitem[Gallegos et~al\mbox{.}(2024)]%
        {gallegos2024bias}
\bibfield{author}{\bibinfo{person}{Isabel~O. Gallegos}, \bibinfo{person}{Ryan~A. Rossi}, \bibinfo{person}{Joe Barrow}, \bibinfo{person}{Md~Mehrab Tanjim}, \bibinfo{person}{Sungchul Kim}, \bibinfo{person}{Franck Dernoncourt}, \bibinfo{person}{Tong Yu}, \bibinfo{person}{Ruiyi Zhang}, {and} \bibinfo{person}{Nesreen~K. Ahmed}.} \bibinfo{year}{2024}\natexlab{}.
\newblock \showarticletitle{Bias and Fairness in Large Language Models: A Survey}.
\newblock \bibinfo{journal}{\emph{Computational Linguistics}} \bibinfo{volume}{50}, \bibinfo{number}{3} (\bibinfo{year}{2024}), \bibinfo{pages}{1097--1179}.
\newblock
\showISSN{0891-2017}


\bibitem[Hartmann et~al\mbox{.}(2023)]%
        {hartmann2023politicalideologyconversationalai}
\bibfield{author}{\bibinfo{person}{Jochen Hartmann}, \bibinfo{person}{Jasper Schwenzow}, {and} \bibinfo{person}{Maximilian Witte}.} \bibinfo{year}{2023}\natexlab{}.
\newblock \bibinfo{title}{The political ideology of conversational AI: Converging evidence on ChatGPT's pro-environmental, left-libertarian orientation}.
\newblock
\newblock
\showeprint[arxiv]{2301.01768}~[cs.CL]
\urldef\tempurl%
\url{https://arxiv.org/abs/2301.01768}
\showURL{%
\tempurl}


\bibitem[Hoes et~al\mbox{.}(2023)]%
        {hoes2023chatgpt}
\bibfield{author}{\bibinfo{person}{Emma Hoes}, \bibinfo{person}{Sacha Altay}, {and} \bibinfo{person}{Juan Bermeo}.} \bibinfo{year}{2023}\natexlab{}.
\newblock \bibinfo{title}{Using ChatGPT to Fight Misinformation: ChatGPT Nails 72\% of 12,000 Verified Claims}.
\newblock
\newblock
\showeprint[psyarxiv]{qnjkf}


\bibitem[Hosseinmardi et~al\mbox{.}(2021)]%
        {hosseinmardi2021examing}
\bibfield{author}{\bibinfo{person}{Homa Hosseinmardi}, \bibinfo{person}{Amir Ghasemian}, \bibinfo{person}{Aaron Clauset}, \bibinfo{person}{Markus Mobius}, \bibinfo{person}{David~M. Rothschild}, {and} \bibinfo{person}{Duncan~J. Watts}.} \bibinfo{year}{2021}\natexlab{}.
\newblock \showarticletitle{Examining the consumption of radical content on YouTube}.
\newblock \bibinfo{journal}{\emph{Proceedings of the National Academy of Sciences}} \bibinfo{volume}{118}, \bibinfo{number}{32} (\bibinfo{year}{2021}), \bibinfo{pages}{e2101967118}.
\newblock


\bibitem[Jakesch et~al\mbox{.}(2023)]%
        {jakesch2023co}
\bibfield{author}{\bibinfo{person}{Maurice Jakesch}, \bibinfo{person}{Advait Bhat}, \bibinfo{person}{Daniel Buschek}, \bibinfo{person}{Lior Zalmanson}, {and} \bibinfo{person}{Mor Naaman}.} \bibinfo{year}{2023}\natexlab{}.
\newblock \showarticletitle{Co-Writing with Opinionated Language Models Affects Users’ Views}. In \bibinfo{booktitle}{\emph{Proceedings of the 2023 CHI Conference on Human Factors in Computing Systems}} (Hamburg, Germany) \emph{(\bibinfo{series}{CHI '23})}. \bibinfo{publisher}{Association for Computing Machinery}, \bibinfo{address}{New York, NY, USA}, Article \bibinfo{articleno}{111}, \bibinfo{numpages}{15}~pages.
\newblock
\showISBNx{9781450394215}


\bibitem[Ji et~al\mbox{.}(2023)]%
        {ji2023survey}
\bibfield{author}{\bibinfo{person}{Ziwei Ji}, \bibinfo{person}{Nayeon Lee}, \bibinfo{person}{Rita Frieske}, \bibinfo{person}{Tiezheng Yu}, \bibinfo{person}{Dan Su}, \bibinfo{person}{Yan Xu}, \bibinfo{person}{Etsuko Ishii}, \bibinfo{person}{Ye~Jin Bang}, \bibinfo{person}{Andrea Madotto}, {and} \bibinfo{person}{Pascale Fung}.} \bibinfo{year}{2023}\natexlab{}.
\newblock \showarticletitle{Survey of hallucination in natural language generation}.
\newblock \bibinfo{journal}{\emph{Comput. Surveys}} \bibinfo{volume}{55}, \bibinfo{number}{12} (\bibinfo{year}{2023}), \bibinfo{pages}{1--38}.
\newblock


\bibitem[Kozyreva et~al\mbox{.}(2024)]%
        {kozyreva2024toolbox}
\bibfield{author}{\bibinfo{person}{Anastasia Kozyreva}, \bibinfo{person}{Philipp Lorenz-Spreen}, \bibinfo{person}{Stefan~M Herzog}, \bibinfo{person}{Ullrich~KH Ecker}, \bibinfo{person}{Stephan Lewandowsky}, \bibinfo{person}{Ralph Hertwig}, \bibinfo{person}{Ayesha Ali}, \bibinfo{person}{Joe Bak-Coleman}, \bibinfo{person}{Sarit Barzilai}, \bibinfo{person}{Melisa Basol}, {et~al\mbox{.}}} \bibinfo{year}{2024}\natexlab{}.
\newblock \showarticletitle{Toolbox of individual-level interventions against online misinformation}.
\newblock \bibinfo{journal}{\emph{Nature Human Behaviour}} \bibinfo{volume}{8}, \bibinfo{number}{6} (\bibinfo{year}{2024}), \bibinfo{pages}{1044--1052}.
\newblock


\bibitem[Lasser et~al\mbox{.}(2022)]%
        {lasser2022social}
\bibfield{author}{\bibinfo{person}{Jana Lasser}, \bibinfo{person}{Segun~Taofeek Aroyehun}, \bibinfo{person}{Almog Simchon}, \bibinfo{person}{Fabio Carrella}, \bibinfo{person}{David Garcia}, {and} \bibinfo{person}{Stephan Lewandowsky}.} \bibinfo{year}{2022}\natexlab{}.
\newblock \bibinfo{title}{Social media sharing by political elites: An asymmetric American exceptionalism}.
\newblock
\newblock
\showeprint[arxiv]{2207.06313}~[cs.CY]
\urldef\tempurl%
\url{https://arxiv.org/abs/2207.06313}
\showURL{%
\tempurl}


\bibitem[Lazer et~al\mbox{.}(2018)]%
        {lazer2018science}
\bibfield{author}{\bibinfo{person}{David~MJ Lazer}, \bibinfo{person}{Matthew~A Baum}, \bibinfo{person}{Yochai Benkler}, \bibinfo{person}{Adam~J Berinsky}, \bibinfo{person}{Kelly~M Greenhill}, \bibinfo{person}{Filippo Menczer}, \bibinfo{person}{Miriam~J Metzger}, \bibinfo{person}{Brendan Nyhan}, \bibinfo{person}{Gordon Pennycook}, \bibinfo{person}{David Rothschild}, {et~al\mbox{.}}} \bibinfo{year}{2018}\natexlab{}.
\newblock \showarticletitle{The science of fake news}.
\newblock \bibinfo{journal}{\emph{Science}} \bibinfo{volume}{359}, \bibinfo{number}{6380} (\bibinfo{year}{2018}), \bibinfo{pages}{1094--1096}.
\newblock


\bibitem[Le~Pochat et~al\mbox{.}(2019)]%
        {pochat2018tranco}
\bibfield{author}{\bibinfo{person}{Victor Le~Pochat}, \bibinfo{person}{Tom Van~Goethem}, \bibinfo{person}{Samaneh Tajalizadehkhoob}, \bibinfo{person}{Maciej Korczynski}, {and} \bibinfo{person}{Wouter Joosen}.} \bibinfo{year}{2019}\natexlab{}.
\newblock \showarticletitle{Tranco: A Research-Oriented Top Sites Ranking Hardened Against Manipulation}. In \bibinfo{booktitle}{\emph{Proceedings 2019 Network and Distributed System Security Symposium}} \emph{(\bibinfo{series}{NDSS 2019})}. \bibinfo{publisher}{Internet Society}, \bibinfo{address}{San Diego, California, USA}, \bibinfo{pages}{1--15}.
\newblock


\bibitem[Li and Sinnamon(2024)]%
        {li2024generativeaisearchengines}
\bibfield{author}{\bibinfo{person}{Alice Li} {and} \bibinfo{person}{Luanne Sinnamon}.} \bibinfo{year}{2024}\natexlab{}.
\newblock \showarticletitle{Generative AI Search Engines as Arbiters of Public Knowledge: An Audit of Bias and Authority}.
\newblock \bibinfo{journal}{\emph{Proceedings of the Association for Information Science and Technology}} \bibinfo{volume}{61}, \bibinfo{number}{1} (\bibinfo{year}{2024}), \bibinfo{pages}{205--217}.
\newblock


\bibitem[Lin et~al\mbox{.}(2023)]%
        {lin2023high}
\bibfield{author}{\bibinfo{person}{Hause Lin}, \bibinfo{person}{Jana Lasser}, \bibinfo{person}{Stephan Lewandowsky}, \bibinfo{person}{Rocky Cole}, \bibinfo{person}{Andrew Gully}, \bibinfo{person}{David~G Rand}, {and} \bibinfo{person}{Gordon Pennycook}.} \bibinfo{year}{2023}\natexlab{}.
\newblock \showarticletitle{{High level of correspondence across different news domain quality rating sets}}.
\newblock \bibinfo{journal}{\emph{PNAS Nexus}} \bibinfo{volume}{2}, \bibinfo{number}{9} (\bibinfo{year}{2023}), \bibinfo{pages}{pgad286}.
\newblock


\bibitem[Liu et~al\mbox{.}(2023)]%
        {liu2023evaluatingverifiabilitygenerativesearch}
\bibfield{author}{\bibinfo{person}{Nelson Liu}, \bibinfo{person}{Tianyi Zhang}, {and} \bibinfo{person}{Percy Liang}.} \bibinfo{year}{2023}\natexlab{}.
\newblock \showarticletitle{Evaluating Verifiability in Generative Search Engines}. In \bibinfo{booktitle}{\emph{Findings of the Association for Computational Linguistics: EMNLP 2023}}. \bibinfo{publisher}{Association for Computational Linguistics}, \bibinfo{address}{Singapore}, \bibinfo{pages}{7001--7025}.
\newblock


\bibitem[Liu et~al\mbox{.}(2024)]%
        {liu2024datasetslargelanguagemodels}
\bibfield{author}{\bibinfo{person}{Yang Liu}, \bibinfo{person}{Jiahuan Cao}, \bibinfo{person}{Chongyu Liu}, \bibinfo{person}{Kai Ding}, {and} \bibinfo{person}{Lianwen Jin}.} \bibinfo{year}{2024}\natexlab{}.
\newblock \bibinfo{title}{Datasets for Large Language Models: A Comprehensive Survey}.
\newblock
\newblock
\showeprint[arxiv]{2402.18041}~[cs.CL]
\urldef\tempurl%
\url{https://arxiv.org/abs/2402.18041}
\showURL{%
\tempurl}


\bibitem[{Llama Team, AI at Meta}(2024)]%
        {dubey2024llama3herdmodels}
\bibfield{author}{\bibinfo{person}{{Llama Team, AI at Meta}}.} \bibinfo{year}{2024}\natexlab{}.
\newblock \bibinfo{title}{The Llama 3 Herd of Models}.
\newblock
\newblock
\showeprint[arxiv]{2407.21783}~[cs.AI]
\urldef\tempurl%
\url{https://arxiv.org/abs/2407.21783}
\showURL{%
\tempurl}


\bibitem[Memon and West(2024)]%
        {memon2024searchenginespostchatgptgenerative}
\bibfield{author}{\bibinfo{person}{Shahan~Ali Memon} {and} \bibinfo{person}{Jevin~D. West}.} \bibinfo{year}{2024}\natexlab{}.
\newblock \bibinfo{title}{Search Engines Post-ChatGPT: How Generative Artificial Intelligence Could Make Search Less Reliable}.
\newblock
\newblock
\showeprint[arxiv]{2402.11707}~[cs.IR]
\urldef\tempurl%
\url{https://arxiv.org/abs/2402.11707}
\showURL{%
\tempurl}


\bibitem[M{\"o}kander et~al\mbox{.}(2023)]%
        {mokander2023auditing}
\bibfield{author}{\bibinfo{person}{Jakob M{\"o}kander}, \bibinfo{person}{Jonas Schuett}, \bibinfo{person}{Hannah~Rose Kirk}, {and} \bibinfo{person}{Luciano Floridi}.} \bibinfo{year}{2023}\natexlab{}.
\newblock \showarticletitle{Auditing large language models: a three-layered approach}.
\newblock \bibinfo{journal}{\emph{AI and Ethics}} \bibinfo{volume}{4}, \bibinfo{number}{4} (\bibinfo{year}{2023}), \bibinfo{pages}{1085--1115}.
\newblock


\bibitem[OpenAI(2024)]%
        {openai2024openaio1card}
\bibfield{author}{\bibinfo{person}{OpenAI}.} \bibinfo{year}{2024}\natexlab{}.
\newblock \bibinfo{title}{OpenAI o1 System Card}.
\newblock
\newblock
\showeprint[arxiv]{2412.16720}~[cs.AI]
\urldef\tempurl%
\url{https://arxiv.org/abs/2412.16720}
\showURL{%
\tempurl}


\bibitem[Ouyang et~al\mbox{.}(2022)]%
        {ouyang2022training}
\bibfield{author}{\bibinfo{person}{Long Ouyang}, \bibinfo{person}{Jeffrey Wu}, \bibinfo{person}{Xu Jiang}, \bibinfo{person}{Diogo Almeida}, \bibinfo{person}{Carroll Wainwright}, \bibinfo{person}{Pamela Mishkin}, \bibinfo{person}{Chong Zhang}, \bibinfo{person}{Sandhini Agarwal}, \bibinfo{person}{Katarina Slama}, \bibinfo{person}{Alex Ray}, {et~al\mbox{.}}} \bibinfo{year}{2022}\natexlab{}.
\newblock \showarticletitle{Training language models to follow instructions with human feedback}.
\newblock \bibinfo{journal}{\emph{Advances in Neural Information Processing Systems}}  \bibinfo{volume}{35} (\bibinfo{year}{2022}), \bibinfo{pages}{27730--27744}.
\newblock


\bibitem[Pennycook and Rand(2019)]%
        {pennycook2019fighting}
\bibfield{author}{\bibinfo{person}{Gordon Pennycook} {and} \bibinfo{person}{David~G. Rand}.} \bibinfo{year}{2019}\natexlab{}.
\newblock \showarticletitle{Fighting misinformation on social media using crowdsourced judgments of news source quality}.
\newblock \bibinfo{journal}{\emph{Proceedings of the National Academy of Sciences}} \bibinfo{volume}{116}, \bibinfo{number}{7} (\bibinfo{year}{2019}), \bibinfo{pages}{2521--2526}.
\newblock


\bibitem[Pfrommer et~al\mbox{.}(2024)]%
        {pfrommer2024rankingmanipulationconversationalsearch}
\bibfield{author}{\bibinfo{person}{Samuel Pfrommer}, \bibinfo{person}{Yatong Bai}, \bibinfo{person}{Tanmay Gautam}, {and} \bibinfo{person}{Somayeh Sojoudi}.} \bibinfo{year}{2024}\natexlab{}.
\newblock \showarticletitle{Ranking Manipulation for Conversational Search Engines}. In \bibinfo{booktitle}{\emph{Proceedings of the 2024 Conference on Empirical Methods in Natural Language Processing}}. \bibinfo{publisher}{Association for Computational Linguistics}, \bibinfo{address}{Miami, Florida, USA}, \bibinfo{pages}{9523--9552}.
\newblock


\bibitem[Qin et~al\mbox{.}(2024)]%
        {qin2024largelanguagemodelseffective}
\bibfield{author}{\bibinfo{person}{Zhen Qin}, \bibinfo{person}{Rolf Jagerman}, \bibinfo{person}{Kai Hui}, \bibinfo{person}{Honglei Zhuang}, \bibinfo{person}{Junru Wu}, \bibinfo{person}{Le Yan}, \bibinfo{person}{Jiaming Shen}, \bibinfo{person}{Tianqi Liu}, \bibinfo{person}{Jialu Liu}, \bibinfo{person}{Donald Metzler}, \bibinfo{person}{Xuanhui Wang}, {and} \bibinfo{person}{Michael Bendersky}.} \bibinfo{year}{2024}\natexlab{}.
\newblock \showarticletitle{Large Language Models are Effective Text Rankers with Pairwise Ranking Prompting}. In \bibinfo{booktitle}{\emph{Findings of the Association for Computational Linguistics: NAACL 2024}}. \bibinfo{publisher}{Association for Computational Linguistics}, \bibinfo{address}{Mexico City, Mexico}, \bibinfo{pages}{1504--1518}.
\newblock


\bibitem[Rahwan et~al\mbox{.}(2019)]%
        {rahwan2019machine}
\bibfield{author}{\bibinfo{person}{Iyad Rahwan}, \bibinfo{person}{Manuel Cebrian}, \bibinfo{person}{Nick Obradovich}, \bibinfo{person}{Josh Bongard}, \bibinfo{person}{Jean-Fran{\c{c}}ois Bonnefon}, \bibinfo{person}{Cynthia Breazeal}, \bibinfo{person}{Jacob~W Crandall}, \bibinfo{person}{Nicholas~A Christakis}, \bibinfo{person}{Iain~D Couzin}, \bibinfo{person}{Matthew~O Jackson}, {et~al\mbox{.}}} \bibinfo{year}{2019}\natexlab{}.
\newblock \showarticletitle{Machine behaviour}.
\newblock \bibinfo{journal}{\emph{Nature}} \bibinfo{volume}{568}, \bibinfo{number}{7753} (\bibinfo{year}{2019}), \bibinfo{pages}{477--486}.
\newblock


\bibitem[Robertson et~al\mbox{.}(2018)]%
        {robertson2018auditingpartisan}
\bibfield{author}{\bibinfo{person}{Ronald~E. Robertson}, \bibinfo{person}{Shan Jiang}, \bibinfo{person}{Kenneth Joseph}, \bibinfo{person}{Lisa Friedland}, \bibinfo{person}{David Lazer}, {and} \bibinfo{person}{Christo Wilson}.} \bibinfo{year}{2018}\natexlab{}.
\newblock \showarticletitle{Auditing Partisan Audience Bias within Google Search}.
\newblock \bibinfo{journal}{\emph{Proceedings of the ACM on human-computer interaction}} \bibinfo{volume}{2}, \bibinfo{number}{CSCW}, Article \bibinfo{articleno}{148} (\bibinfo{year}{2018}), \bibinfo{numpages}{22}~pages.
\newblock


\bibitem[Salinas et~al\mbox{.}(2023)]%
        {salinas2023imracistbutdiscovering}
\bibfield{author}{\bibinfo{person}{Abel Salinas}, \bibinfo{person}{Louis Penafiel}, \bibinfo{person}{Robert McCormack}, {and} \bibinfo{person}{Fred Morstatter}.} \bibinfo{year}{2023}\natexlab{}.
\newblock \bibinfo{title}{"Im not Racist but...": Discovering Bias in the Internal Knowledge of Large Language Models}.
\newblock
\newblock
\showeprint[arxiv]{2310.08780}~[cs.CL]
\urldef\tempurl%
\url{https://arxiv.org/abs/2310.08780}
\showURL{%
\tempurl}


\bibitem[Santurkar et~al\mbox{.}(2023)]%
        {santurkar2023whose}
\bibfield{author}{\bibinfo{person}{Shibani Santurkar}, \bibinfo{person}{Esin Durmus}, \bibinfo{person}{Faisal Ladhak}, \bibinfo{person}{Cinoo Lee}, \bibinfo{person}{Percy Liang}, {and} \bibinfo{person}{Tatsunori Hashimoto}.} \bibinfo{year}{2023}\natexlab{}.
\newblock \showarticletitle{Whose Opinions Do Language Models Reflect?}. In \bibinfo{booktitle}{\emph{Proceedings of the 40th International Conference on Machine Learning}} \emph{(\bibinfo{series}{Proceedings of Machine Learning Research}, Vol.~\bibinfo{volume}{202})}. \bibinfo{publisher}{PMLR}, \bibinfo{address}{Honolulu, Hawaii, USA}, \bibinfo{pages}{29971--30004}.
\newblock


\bibitem[Schlichtkrull(2024)]%
        {schlichtkrull2024generatingmediabackgroundchecks}
\bibfield{author}{\bibinfo{person}{Michael Schlichtkrull}.} \bibinfo{year}{2024}\natexlab{}.
\newblock \bibinfo{title}{Generating Media Background Checks for Automated Source Critical Reasoning}.
\newblock
\newblock
\showeprint[arxiv]{2409.00781}~[cs.CL]
\urldef\tempurl%
\url{https://arxiv.org/abs/2409.00781}
\showURL{%
\tempurl}


\bibitem[Sharma et~al\mbox{.}(2024)]%
        {sharma2024generative}
\bibfield{author}{\bibinfo{person}{Nikhil Sharma}, \bibinfo{person}{Q.~Vera Liao}, {and} \bibinfo{person}{Ziang Xiao}.} \bibinfo{year}{2024}\natexlab{}.
\newblock \showarticletitle{Generative Echo Chamber? Effect of LLM-Powered Search Systems on Diverse Information Seeking}. In \bibinfo{booktitle}{\emph{Proceedings of the CHI Conference on Human Factors in Computing Systems}} (Honolulu, HI, USA) \emph{(\bibinfo{series}{CHI '24})}. \bibinfo{publisher}{Association for Computing Machinery}, \bibinfo{address}{New York, NY, USA}, Article \bibinfo{articleno}{1033}, \bibinfo{numpages}{17}~pages.
\newblock
\showISBNx{9798400703300}


\bibitem[Simchon et~al\mbox{.}(2024)]%
        {simchon2024persuasive}
\bibfield{author}{\bibinfo{person}{Almog Simchon}, \bibinfo{person}{Matthew Edwards}, {and} \bibinfo{person}{Stephan Lewandowsky}.} \bibinfo{year}{2024}\natexlab{}.
\newblock \showarticletitle{{The persuasive effects of political microtargeting in the age of generative artificial intelligence}}.
\newblock \bibinfo{journal}{\emph{PNAS Nexus}} \bibinfo{volume}{3}, \bibinfo{number}{2} (\bibinfo{date}{01} \bibinfo{year}{2024}), \bibinfo{pages}{pgae035}.
\newblock
\showISSN{2752-6542}


\bibitem[Simmons(2023)]%
        {simmons2023moralmimicrylargelanguage}
\bibfield{author}{\bibinfo{person}{Gabriel Simmons}.} \bibinfo{year}{2023}\natexlab{}.
\newblock \showarticletitle{Moral Mimicry: Large Language Models Produce Moral Rationalizations Tailored to Political Identity}. In \bibinfo{booktitle}{\emph{Proceedings of the 61st Annual Meeting of the Association for Computational Linguistics}}. \bibinfo{publisher}{Association for Computational Linguistics}, \bibinfo{address}{Toronto, Canada}, \bibinfo{pages}{282--297}.
\newblock


\bibitem[Spatharioti et~al\mbox{.}(2023)]%
        {spatharioti2023comparingtraditionalllmbasedsearch}
\bibfield{author}{\bibinfo{person}{Sofia~Eleni Spatharioti}, \bibinfo{person}{David~M. Rothschild}, \bibinfo{person}{Daniel~G. Goldstein}, {and} \bibinfo{person}{Jake~M. Hofman}.} \bibinfo{year}{2023}\natexlab{}.
\newblock \bibinfo{title}{Comparing Traditional and LLM-based Search for Consumer Choice: A Randomized Experiment}.
\newblock
\newblock
\showeprint[arxiv]{2307.03744}~[cs.HC]
\urldef\tempurl%
\url{https://arxiv.org/abs/2307.03744}
\showURL{%
\tempurl}


\bibitem[Sun et~al\mbox{.}(2023)]%
        {sun2023chatgptgoodsearchinvestigating}
\bibfield{author}{\bibinfo{person}{Weiwei Sun}, \bibinfo{person}{Lingyong Yan}, \bibinfo{person}{Xinyu Ma}, \bibinfo{person}{Shuaiqiang Wang}, \bibinfo{person}{Pengjie Ren}, \bibinfo{person}{Zhumin Chen}, \bibinfo{person}{Dawei Yin}, {and} \bibinfo{person}{Zhaochun Ren}.} \bibinfo{year}{2023}\natexlab{}.
\newblock \showarticletitle{Is {C}hat{GPT} Good at Search? Investigating Large Language Models as Re-Ranking Agents}. In \bibinfo{booktitle}{\emph{Proceedings of the 2023 Conference on Empirical Methods in Natural Language Processing}}. \bibinfo{publisher}{Association for Computational Linguistics}, \bibinfo{address}{Singapore}, \bibinfo{pages}{14918--14937}.
\newblock


\bibitem[Suri et~al\mbox{.}(2024)]%
        {suri2024usegenerativesearchengines}
\bibfield{author}{\bibinfo{person}{Siddharth Suri}, \bibinfo{person}{Scott Counts}, \bibinfo{person}{Leijie Wang}, \bibinfo{person}{Chacha Chen}, \bibinfo{person}{Mengting Wan}, \bibinfo{person}{Tara Safavi}, \bibinfo{person}{Jennifer Neville}, \bibinfo{person}{Chirag Shah}, \bibinfo{person}{Ryen~W. White}, \bibinfo{person}{Reid Andersen}, \bibinfo{person}{Georg Buscher}, \bibinfo{person}{Sathish Manivannan}, \bibinfo{person}{Nagu Rangan}, {and} \bibinfo{person}{Longqi Yang}.} \bibinfo{year}{2024}\natexlab{}.
\newblock \bibinfo{title}{The Use of Generative Search Engines for Knowledge Work and Complex Tasks}.
\newblock
\newblock
\showeprint[arxiv]{2404.04268}~[cs.IR]
\urldef\tempurl%
\url{https://arxiv.org/abs/2404.04268}
\showURL{%
\tempurl}


\bibitem[Trielli and Diakopoulos(2019)]%
        {trielli2019search}
\bibfield{author}{\bibinfo{person}{Daniel Trielli} {and} \bibinfo{person}{Nicholas Diakopoulos}.} \bibinfo{year}{2019}\natexlab{}.
\newblock \showarticletitle{Search as News Curator: The Role of Google in Shaping Attention to News Information}. In \bibinfo{booktitle}{\emph{Proceedings of the 2019 CHI Conference on Human Factors in Computing Systems}} (Glasgow, Scotland Uk) \emph{(\bibinfo{series}{CHI '19})}. \bibinfo{publisher}{Association for Computing Machinery}, \bibinfo{address}{New York, NY, USA}, \bibinfo{pages}{1–15}.
\newblock
\showISBNx{9781450359702}


\bibitem[Urman and Makhortykh(2025)]%
        {urman2023silence}
\bibfield{author}{\bibinfo{person}{Aleksandra Urman} {and} \bibinfo{person}{Mykola Makhortykh}.} \bibinfo{year}{2025}\natexlab{}.
\newblock \showarticletitle{The silence of the LLMs: Cross-lingual analysis of guardrail-related political bias and false information prevalence in ChatGPT, Google Bard (Gemini), and Bing Chat}.
\newblock \bibinfo{journal}{\emph{Telematics and Informatics}}  \bibinfo{volume}{96} (\bibinfo{year}{2025}), \bibinfo{pages}{102211}.
\newblock
\showISSN{0736-5853}


\bibitem[Vu et~al\mbox{.}(2023)]%
        {vu2023freshllmsrefreshinglargelanguage}
\bibfield{author}{\bibinfo{person}{Tu Vu}, \bibinfo{person}{Mohit Iyyer}, \bibinfo{person}{Xuezhi Wang}, \bibinfo{person}{Noah Constant}, \bibinfo{person}{Jerry Wei}, \bibinfo{person}{Jason Wei}, \bibinfo{person}{Chris Tar}, \bibinfo{person}{Yun-Hsuan Sung}, \bibinfo{person}{Denny Zhou}, \bibinfo{person}{Quoc Le}, {and} \bibinfo{person}{Thang Luong}.} \bibinfo{year}{2023}\natexlab{}.
\newblock \bibinfo{title}{FreshLLMs: Refreshing Large Language Models with Search Engine Augmentation}.
\newblock
\newblock
\showeprint[arxiv]{2310.03214}~[cs.CL]
\urldef\tempurl%
\url{https://arxiv.org/abs/2310.03214}
\showURL{%
\tempurl}


\bibitem[Williams et~al\mbox{.}(2024)]%
        {williams2024bridgingsocialmediasearch}
\bibfield{author}{\bibinfo{person}{Evan~M. Williams}, \bibinfo{person}{Peter Carragher}, {and} \bibinfo{person}{Kathleen~M. Carley}.} \bibinfo{year}{2024}\natexlab{}.
\newblock \bibinfo{title}{Bridging Social Media and Search Engines: Dredge Words and the Detection of Unreliable Domains}.
\newblock
\newblock
\showeprint[arxiv]{2406.11423}~[cs.SI]
\urldef\tempurl%
\url{https://arxiv.org/abs/2406.11423}
\showURL{%
\tempurl}


\bibitem[Wu et~al\mbox{.}(2020)]%
        {wu2020providing}
\bibfield{author}{\bibinfo{person}{Zhijing Wu}, \bibinfo{person}{Mark Sanderson}, \bibinfo{person}{B.~Barla Cambazoglu}, \bibinfo{person}{W.~Bruce Croft}, {and} \bibinfo{person}{Falk Scholer}.} \bibinfo{year}{2020}\natexlab{}.
\newblock \showarticletitle{Providing Direct Answers in Search Results: A Study of User Behavior}. In \bibinfo{booktitle}{\emph{Proceedings of the 29th ACM International Conference on Information \& Knowledge Management}} (Virtual Event, Ireland) \emph{(\bibinfo{series}{CIKM '20})}. \bibinfo{publisher}{Association for Computing Machinery}, \bibinfo{address}{New York, NY, USA}, \bibinfo{pages}{1635–1644}.
\newblock
\showISBNx{9781450368599}


\bibitem[Xiong et~al\mbox{.}(2024)]%
        {xiong2024searchengineservicesmeet}
\bibfield{author}{\bibinfo{person}{Haoyi Xiong}, \bibinfo{person}{Jiang Bian}, \bibinfo{person}{Yuchen Li}, \bibinfo{person}{Xuhong Li}, \bibinfo{person}{Mengnan Du}, \bibinfo{person}{Shuaiqiang Wang}, \bibinfo{person}{Dawei Yin}, {and} \bibinfo{person}{Sumi Helal}.} \bibinfo{year}{2024}\natexlab{}.
\newblock \showarticletitle{When Search Engine Services Meet Large Language Models: Visions and Challenges}.
\newblock \bibinfo{journal}{\emph{IEEE Transactions on Services Computing}} \bibinfo{volume}{17}, \bibinfo{number}{6} (\bibinfo{year}{2024}), \bibinfo{pages}{4558--4577}.
\newblock


\end{thebibliography}

\appendix

\section{Complete Prompts}

In the default configuration, the following prompts are used in our experiments.

System prompt:

\prompt{System prompt}{
You are an assistant to determine the credibility of websites.
}

Task instruction:

\prompt{Task instruction}{
Rate the website's credibility: \{domain\}, on a scale between 0 and 1, where 0 means very low credibility and 1 means very high credibility. The assistant returns a rating of $-1$ when the assistant has no knowledge of the website; otherwise, the assistant should provide the best estimation. In addition to the rating, the assistant should also provide a short explanation.
The output should be in JSON format and follow the following schema. Only return the JSON object, nothing else.
\{`url': `example.com', `rating': 0.5, `explanation': `The example website is known to post credible content.'\}
}

Note that we specifically instruct the models to return the results in JSON format for easy parsing.
We use the JSON mode for GPT and Gemini models to ensure their LLM responses are in valid JSON format.
Unfortunately, the Llama models hosted by \url{together.ai} do not support JSON mode.
They are usually able to follow the instructions and return the responses in JSON format, but extra cleaning and re-formatting are sometimes required to correctly parse the responses.

When testing the impact of partisan roles, we modify the system prompt:

\prompt{Partisan role assignment}{
You are an assistant to determine the credibility of websites.
You identify as \{role\} on the US political spectrum.
}

In the experiments, \{role\} is replaced with the one of ``a Democrat,'' ``an Independent,'' or ``a Republican.''
The task instruction remains the same.

\section{Response Examples}

In this section, we provide some response examples from the LLMs.
Table~\ref{tab:reuters} lists the LLM ratings of \url{reuters.com}.
Table~\ref{tab:aldf} lists the LLM ratings of \url{aldf.com}.

\begin{table*}
    \centering
    \caption{Ratings of \url{reuters.com} by different LLMs.}
    \begin{tabular}{l|l|p{13cm}}
    \hline
    Model & Rating & Explanation \\
    \hline
    GPT-4o mini & 1 & Reuters is a well-established international news organization known for its commitment to factual reporting and journalistic integrity. It has a strong reputation for accuracy and is widely regarded as a reliable source of news. \\
    \hline
    GPT-4o & 1 & Reuters is a well-established international news organization known for its accurate and unbiased reporting. It is widely respected in the journalism community and has a long history of providing reliable news. \\
    \hline
    GPT-4 & 0.95 & Reuters is a well-established and highly reputable global news organization known for its commitment to accurate and unbiased reporting. \\
    \hline
    o1-mini & 0.9 & Reuters is a well-established news organization known for reliable and unbiased reporting. \\
    \hline
    Gemini 1.5 Flash & 0.9 & Reuters is a well-established and reputable news agency known for its journalistic integrity and fact-checking practices. \\
    \hline
    Gemini 1.5 Pro & 1 & Reuters is a highly reputable international news agency known for its journalistic integrity and factual reporting. \\
    \hline
    Llama 3.1 8B & 0.9 & Reuters is a well-established and reputable news agency with a long history of providing accurate and unbiased reporting. \\
    \hline
    Llama 3.1 70B & 0.95 & Reuters is a well-established and reputable news agency with a long history of providing accurate and unbiased reporting. It is widely considered a trusted source of news and information. \\
    \hline
    Llama 3.1 405B & 0.9 & Reuters is a well-established and reputable international news agency with a long history of providing accurate and unbiased reporting. \\
    \hline
    \end{tabular}
    \label{tab:reuters}
\end{table*}

\begin{table*}
    \centering
    \caption{Ratings of \url{aldf.com} by different LLMs. Five models believe the website belongs to the Animal Legal Defense Fund (\url{aldf.org}).}
    \begin{tabular}{l|l|p{13cm}}
    \hline
    Model & Rating & Explanation \\
    \hline
    GPT-4o mini & 0.8 & The Animal Legal Defense Fund (ALDF) is a well-established organization focused on animal rights and legal advocacy. It is recognized for its credible efforts in promoting animal welfare and has a strong reputation in the field. \\
    \hline
    GPT-4o & -1 & The assistant has no knowledge of the website aldf.com. \\
    \hline
    GPT-4 & -1 & I do not have enough information to assess the credibility of aldf.com. \\
    \hline
    o1-mini & 0.7 & The website aldf.com is associated with the American Lebanese Democratic Federation, a recognized organization, indicating a moderate to high level of credibility. \\
    \hline
    Gemini 1.5 Flash & -1 & I have no information about this website. \\
    \hline
    Gemini 1.5 Pro & 1 & ALDF.org is a highly reputable website of the Animal Legal Defense Fund, a well-known non-profit organization. \\
    \hline
    Llama 3.1 8B & 0.8 & The Animal Legal Defense Fund (ALDF) is a well-established and reputable organization that provides accurate and reliable information on animal law and advocacy. \\
    \hline
    Llama 3.1 70B & 0.8 & The Animal Legal Defense Fund (ALDF) is a well-established and reputable organization that advocates for animal rights and welfare. Their website provides credible information on animal law and related topics. \\
    \hline
    Llama 3.1 405B & 0.8 & The Animal Legal Defense Fund (aldf.org) is a well-established organization with a clear mission and a history of advocating for animal welfare through the legal system. Their website provides credible information on animal law and related topics. \\
    \hline
    \end{tabular}
    \label{tab:aldf}
\end{table*}

\section{Additional Results}

\begin{figure}
    \centering
    \includegraphics[width=0.9\columnwidth]{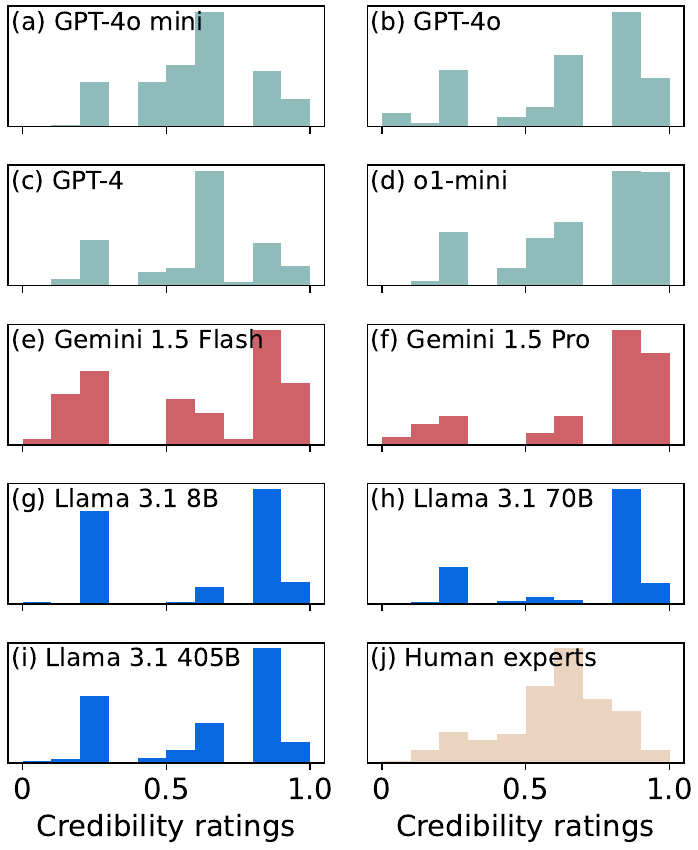}
    \caption{
    Distribution of domain credibility ratings from different LLMs and human experts.
    }
    \Description{Distribution of domain credibility ratings from different LLMs and human experts.}
    \label{fig:distribution}
\end{figure}

In Figure~\ref{fig:distribution}, we show the distribution of domain credibility ratings from different LLMs and human experts.

\end{document}